\documentclass[review]{elsarticle}

\usepackage{rotating}
\usepackage{subfiles}
\usepackage{subcaption}
\usepackage{graphicx}
\usepackage{multirow}
\usepackage{adjustbox}		
\usepackage{array}			
\usepackage{pifont}
\usepackage[table]{xcolor}	
\usepackage{booktabs}		

\usepackage{hyperref}

\newcolumntype{R}[2]{%
    >{\adjustbox{angle=#1,lap=\width-(#2)}\bgroup}%
    l%
    <{\egroup}%
}
\newcommand*\rot{\multicolumn{1}{R{45}{1em}}}

\newcommand{\n}{$\times$}
\newcommand{\y}{\ding{52}}

\definecolor{veryLightGray}{RGB}{230, 230, 230}

\begin{document}
	
	\begin{frontmatter}
		
		\title{Moving Objects Detection with a Moving Camera: A Comprehensive Review}
		
		\author[mymainaddress]{Marie-Neige Chapel}
		\author[mysecondaryaddress]{Thierry Bouwmans}
		
		\address[mymainaddress]{Lab. L3I, LRUniv., Avenue Albert Einstein, 17000 La Rochelle, France}
		\address[mysecondaryaddress]{Lab. MIA, LRUniv., Avenue Albert Einstein, 17000 La Rochelle, France}

		\begin{abstract}
			During about 30 years, a lot of research teams have worked on the big challenge of detection of moving objects in various challenging environments. First applications concern static cameras but with the rise of the mobile sensors studies on moving cameras have emerged over time. In this survey, we propose to identify and categorize the different existing methods found in the literature. For this purpose, we propose to classify these methods according to the choose of the scene representation: one plane or several parts. Inside these two categories, the methods are grouped according to eight different approaches: panoramic background subtraction, dual cameras, motion compensation, subspace segmentation, motion segmentation, plane+parallax, multi planes and split image in blocks. A reminder of methods for static cameras is provided as well as the challenges with both static and moving cameras. Publicly available datasets and evaluation metrics are also surveyed in this paper.
		\end{abstract}
		
		
	\end{frontmatter}
	

%

\section{Introduction}
Cameras are more and more present in our daily lives whether it is in the streets, in our homes and even in our pockets with smart-phones. Many real applications \cite{P0C0-Survey-290} are based on videos taken either by static or moving cameras such as in video surveillance of human activities \cite{P0C0-A-1}, visual observation of animals \cite{P0C0-A-79-1,P0C0-A-79-1-1,P0C0-A-79-1-2}, home care \cite{P0C0-A-1}, optical motion capture \cite{P0C0-A-100} and multimedia applications \cite{P0C0-A-200}. In their process, these applications often require a moving objects detection step followed by tracking and recognition steps. Since 30 years, moving objects detection is thus surely among the most investigated field in computer vision providing a big amount of publications. First, methods were developed for static cameras but, in the last two decades with the expansion of sensors, approaches with moving cameras have been of many interests giving more challenging situations to handle. However, many challenges have been identified in the literature and are related either to the cameras, to the background or to the moving objects of the filmed scenes. 

A lot of surveys in the literature are about moving objects detection in the case of static cameras. In 2000, Mc Ivor \cite{P0C0-Survey-1} surveyed nine algorithms allowing a first comparison of the models. However, this survey is mainly limited on a description of the algorithms. In 2004, Piccardi \cite{P0C0-Survey-2} provided a review on seven methods and an original categorization based on speed, memory requirements and accuracy. This review allows the readers to compare the complexity of the different methods and effectively helps them to select the most adapted method for their specific application. In 2005, Cheung and Kamath \cite{P0C0-Survey-3} classified several methods into non-recursive and recursive techniques. Following this classification, Elhabian et al.\cite{P0C0-Survey-4} provided a large survey in background modeling. However, this classification in terms of non-recursive and recursive techniques is more suitable for the background maintenance scheme than for the background modeling one. In their review in 2010, Cristiani et al. \cite{P0C0-Survey-5} distinguished the most popular background subtraction algorithms by means of their sensor utilization: single monocular sensor or multiple sensors. In 2014, Elgammal \cite{P0C0-Survey-5} provided a chapter on background subtraction for static and moving cameras over 120 papers. Since 2008, Bouwmans et al. \cite{P0C0-Survey-10} initiated several comprehensive surveys classifying each approaches following the employed models that can be classified into the following main chronological categories: traditional models, recent models and prospective models that employed both mathematical, machine learning and signal processing models. These different surveys concern either all the categories \cite{P0C0-Survey-13,P0C0-Survey-14,P0C0-Survey-24}, sub-categories (i.e. statistical models \cite{P0C0-Survey-10}, fuzzy models \cite{P0C0-Survey-11}, decomposition into low-rank plus additive matrices \cite{P0C0-Survey-25} or part of sub-categories (i.e. Mixture of Gaussian models (GMM) \cite{P0C0-Survey-20}, subspace learning models \cite{P0C0-Survey-21}, Robust Principal Component Analysis (RPCA) models \cite{P0C0-Survey-23}, dynamic RCPA models \cite{P3C1-PCP-1030}, and deep learning models \cite{P0C0-Survey-27}).

Sometimes, these previous surveys presented in a sub-part extensions of background subtraction methods to static cameras for moving cameras. One can also find sub-parts that concern moving cameras in object tracking and surveillance surveys \cite{Moeslund2001,Yilmaz2006,Cristani2010,Joshi2012}. However, the techniques addressing the case of moving cameras are more and more numerous and can be the target of whole study as proven by recent reviews \cite{P0C0-Survey-5,Komagal2018,Yazdi2018}. In 2014, Elgammal \cite{P0C0-Survey-5} give an entire chapter on background subtraction techniques for moving camera classifying them into traditional and recent methods. In 2018, Komagal and Yogameena \cite{Komagal2018} chose to review foreground segmentation approaches with a Pan Tilt Zoom (PTZ) camera but those techniques cannot usually be employed with freely moving cameras. In 2018, Yazdi and Bouwmans \cite{Yazdi2018} presented the most complete survey on the subject, to the best of our knowledge. The methods are presented according to challenges and a classification into four broad categories are employed but the review suffers from a lack of completeness. Thus, there is a need of a full comprehensive survey for moving objects detection with moving cameras.

In this context, we propose to fully review methods about moving objects detection with a moving camera. The aim is thus to present a review of the traditional and recent techniques used by categorizing them and making the assessment of the methods regarding the challenges. It is dedicated for students, engineers, young researchers and confirmed researchers in the field. It could serve as basis for courses too and considered as the reference in the field. The paper is organized as follows. First, we define notions of moving objects and moving cameras in Section~\ref{sec:definitions} in order to delimit the scope of this survey. Second, we investigate the different challenges met in videos taken by static and moving cameras in Section \ref{sec:Challenges}. In Section~\ref{sec:background_subtraction_with_a_stationary_camera}, we carefully present the general process of background subtraction method with a static camera by providing a background knowledge to well understand extensions of background subtraction methods in the case of moving cameras. In Section ~\ref{sec:methods_classification}, we provide an original classification of the methods about moving objects detection with a moving camera. Then, evaluation metrics and publicly available datasets are presented in Section~\ref{sec:datasets_and_evaluation_metrics}. Finally, we conclude the paper by a discussion and perspectives for future work.

%
\section{Preliminaries}
\label{sec:definitions}
In this section, we clearly state notions of moving objects and moving cameras that defined the kind of methods that are reviewed in this paper.

\subsection{Moving objects}
\label{sec:definition_moving_object}

In physics, a motion is described by a change in position of an object over time according to a frame of reference attached to an observer. In our case, the observer is the camera and we will describe the observations for two kind of cameras: stationary and moving. For a stationary camera, the background appears static in the video stream of the camera and a moving object appears moving. Displacements of an object in the scene is called the \emph{local motion}. In the case of a moving camera, both of them appear moving. The background appears moving because of the \emph{global motion} and the distinction between a moving object and the static scene is complicated.

The range of moving objects is large, ranging from pedestrians to waving trees. But among of these objects, only a subpart has to be labeled as moving. The objects like waving trees, ocean waves or escalators are part of so-called \emph{dynamic background} and have to be labeled as background. Conversely, pedestrians, cars or animals are objects with "significantly" motions and the subjects of applications about which we interest in this paper.

An object can be represented in many different ways \cite{Yilmaz2006}. In this survey, we are going to see that two kinds of representation are generally used for the moving object detection: a bounding box or a silhouette. The bounding box is usually used in tracking methods where only a rough region of the moving object is needed. The bounding box contains pixels from the background and from the moving object. Conversely, a silhouette provides accuracy information on the moving object position since every pixel in the silhouette has to belong to the object. Silhouette results are needed for some applications like motion capture.

\subsection{Moving cameras}
The specificity of a moving camera compared to a static one, is that a static object appears moving in the video stream. This motion is caused by the motion of the camera also called the \emph{ego motion}. As well as a moving object, the physics definition of motion can be applied to a camera. In addition to displacements in the 3D space, the camera can also perform rotations, named pan, tilt and roll.

Among moving cameras, there are two types of cameras: freely moving camera and constrained moving camera. As its name suggest, freely moving camera performs any kind of motion without any constraint. This camera is hand-held camera, smartphone or drone. In the category of constrained cameras, the most famous example is the PTZ camera. This camera can only perform rotations since its optical center is fixed. Even if this camera doesn't change in position, rotations are enough to defined it as a moving camera.
%

\section{Challenges}
\label{sec:Challenges}

\begin{figure*}
	\begin{minipage}{\textwidth}
		\centering
		\includegraphics[width=0.2\textwidth]{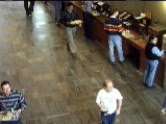}
		\includegraphics[width=0.2\textwidth]{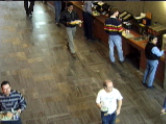}
		\includegraphics[width=0.2\textwidth]{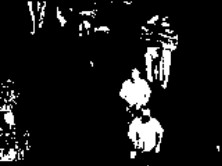}
		\subcaption{Bootstrapping}
	\end{minipage}
	\begin{minipage}{\textwidth}
		\centering
		\includegraphics[width=0.2\textwidth]{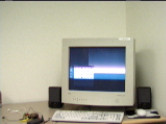}
		\includegraphics[width=0.2\textwidth]{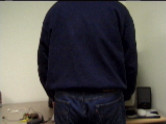}
		\includegraphics[width=0.2\textwidth]{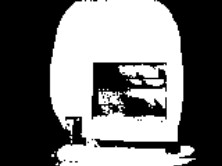}
		\subcaption{Camouflage}
	\end{minipage}
	\begin{minipage}{\textwidth}
		\centering
		\includegraphics[width=0.2\textwidth]{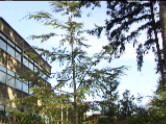}
		\includegraphics[width=0.2\textwidth]{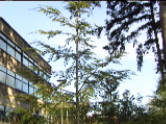}
		\includegraphics[width=0.2\textwidth]{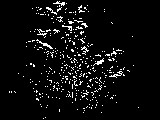}
		\subcaption{Dynamic background}
	\end{minipage}
	\begin{minipage}{\textwidth}
		\centering
		\includegraphics[width=0.2\textwidth]{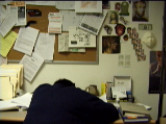}
		\includegraphics[width=0.2\textwidth]{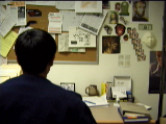}
		\includegraphics[width=0.2\textwidth]{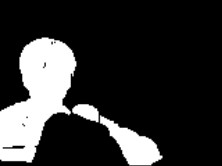}
		\subcaption{Foreground aperture}
	\end{minipage}
	\begin{minipage}{\textwidth}
		\centering
		\includegraphics[width=0.2\textwidth]{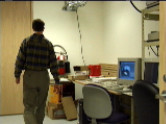}
		\includegraphics[width=0.2\textwidth]{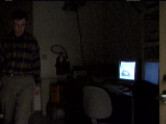}
		\includegraphics[width=0.2\textwidth]{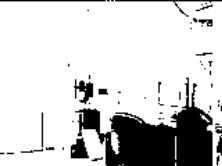}
		\subcaption{Illumination changes}
	\end{minipage}
	\begin{minipage}{\textwidth}
		\centering
		\includegraphics[width=0.2\textwidth]{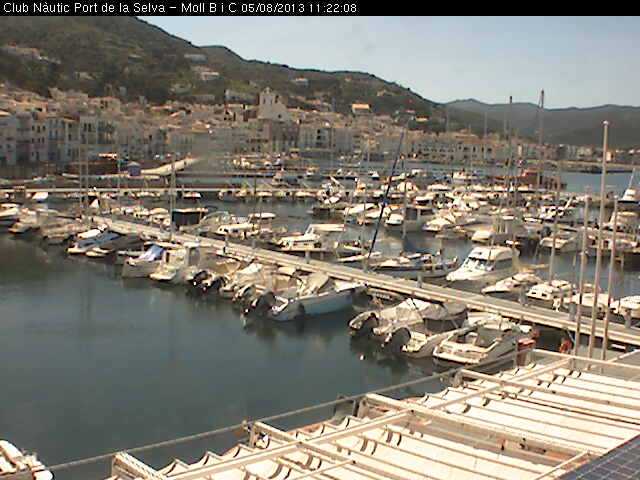}
		\includegraphics[width=0.2\textwidth]{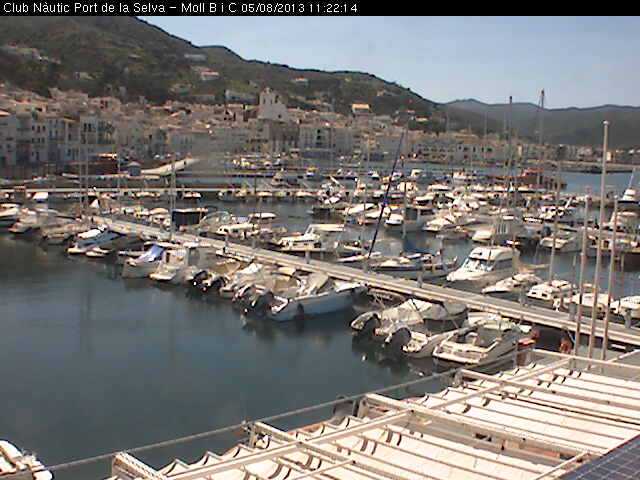}
		\includegraphics[width=0.2\textwidth]{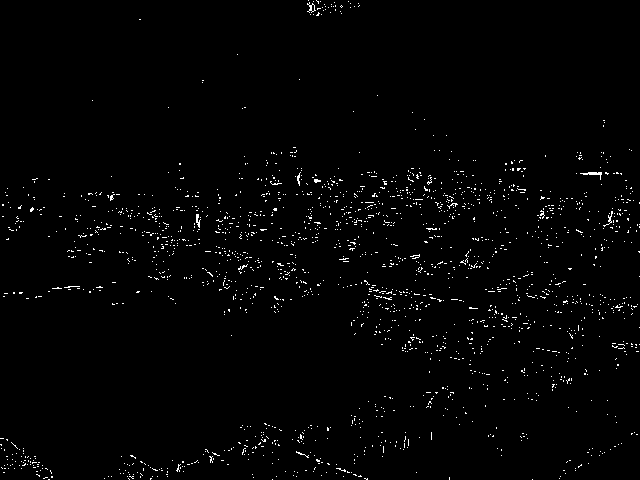}
		\subcaption{Low frame rate}
	\end{minipage}
	\caption{Illustrations of background subtraction challenges. Images come from the Wallflower (Bootstrap, Camouflage, ForegroundAperture, WavingTrees, LightSwitch sequences) dataset and the ChangeDetection.net (port\_0\_17fps sequence) dataset. The last column is the result of Gaussian mixture-based background/foreground segmentation in the OpenCV library.}
	\label{fig:bs_challenges_1}
\end{figure*}

\begin{figure*}
	\begin{minipage}{\textwidth}
		\centering
		\includegraphics[width=0.2\textwidth]{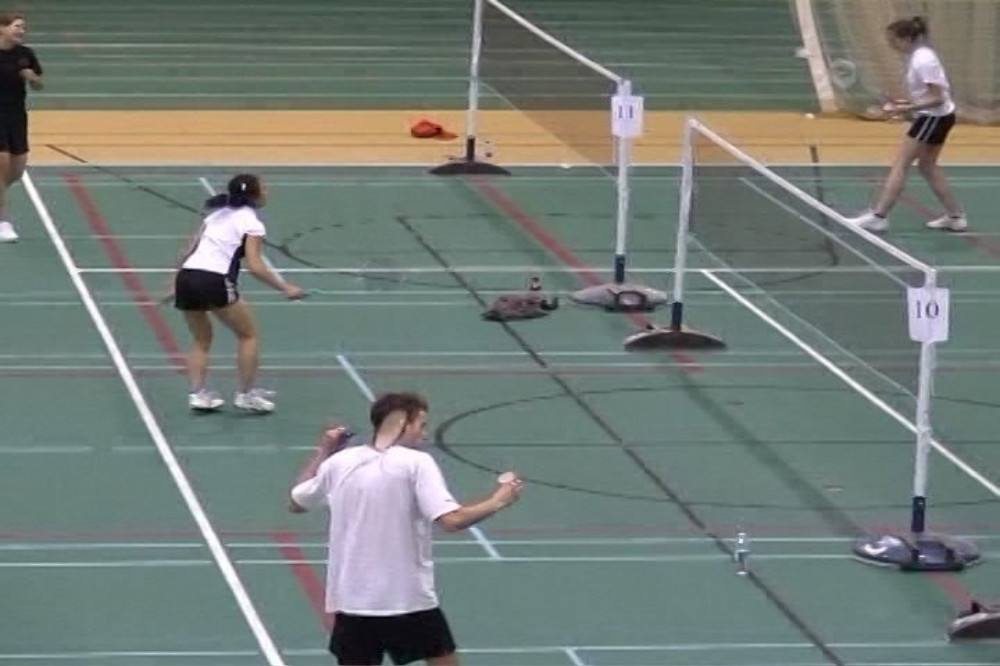}
		\includegraphics[width=0.2\textwidth]{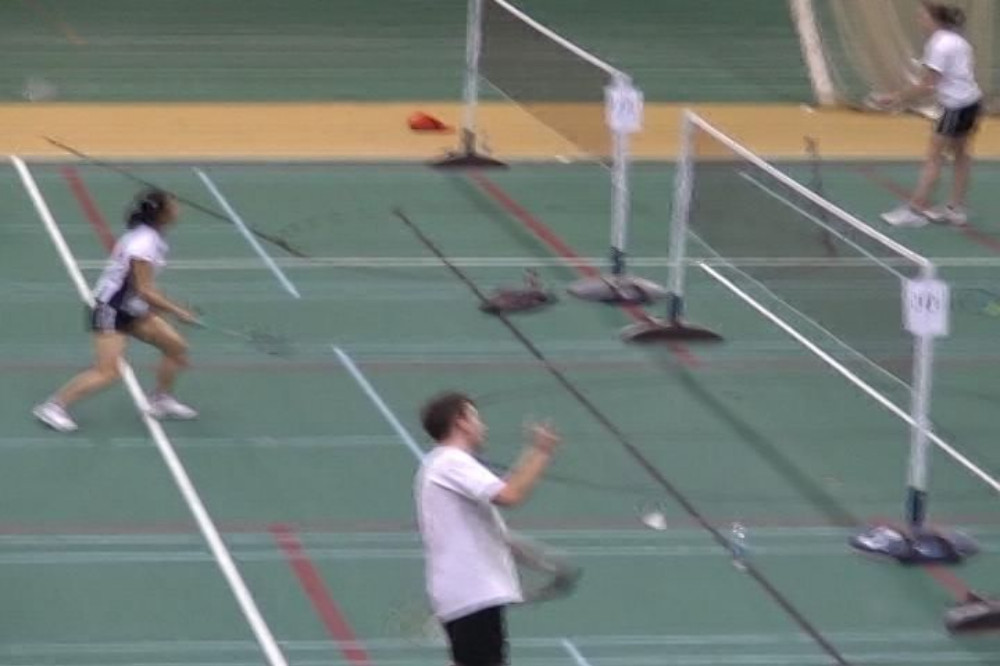}
		\includegraphics[width=0.2\textwidth]{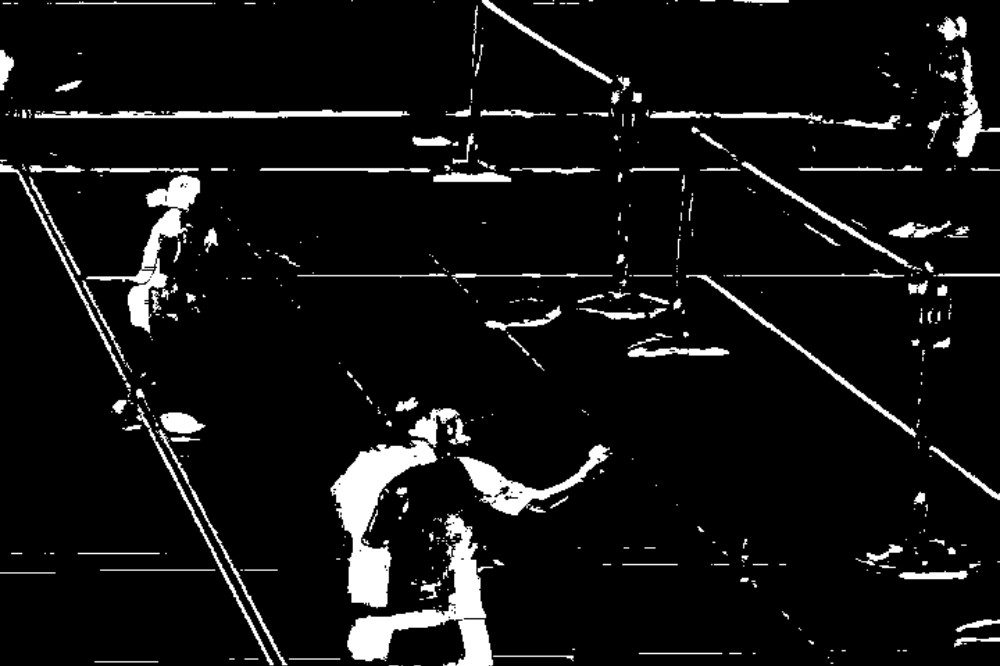}
		\subcaption{Motion blur}
	\end{minipage}
	\begin{minipage}{\textwidth}
		\centering
		\includegraphics[width=0.2\textwidth]{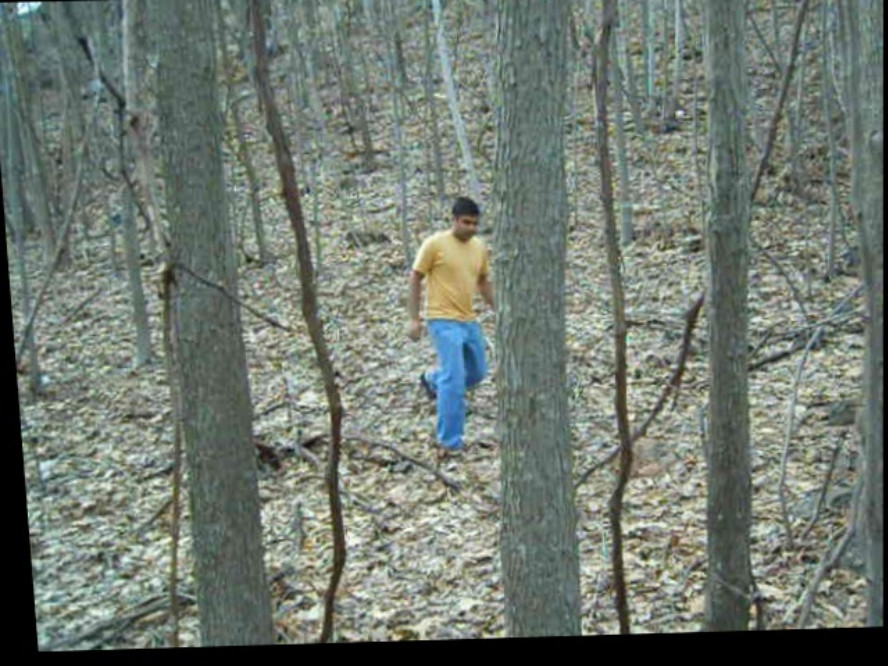}
		\includegraphics[width=0.2\textwidth]{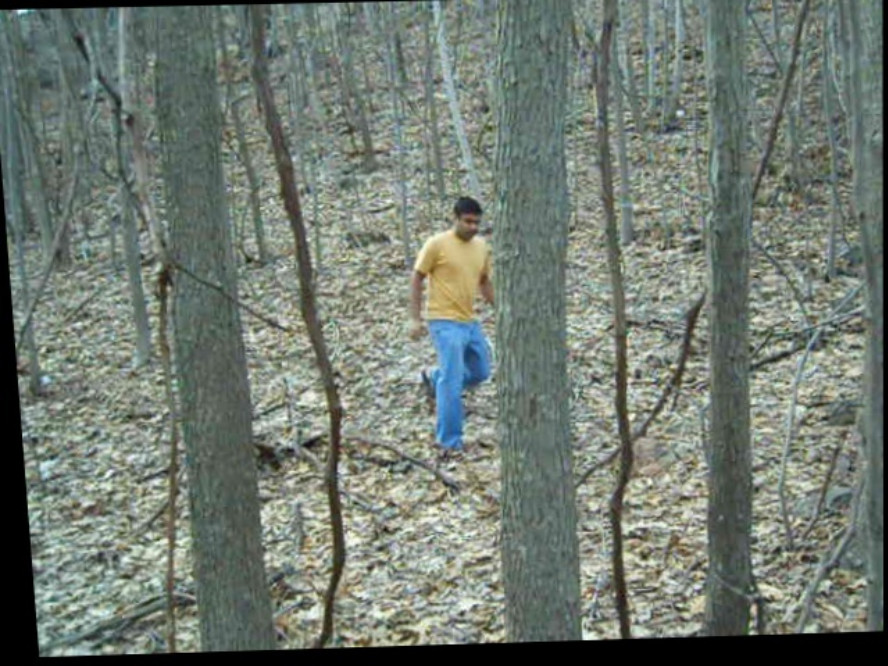}
		\includegraphics[width=0.2\textwidth]{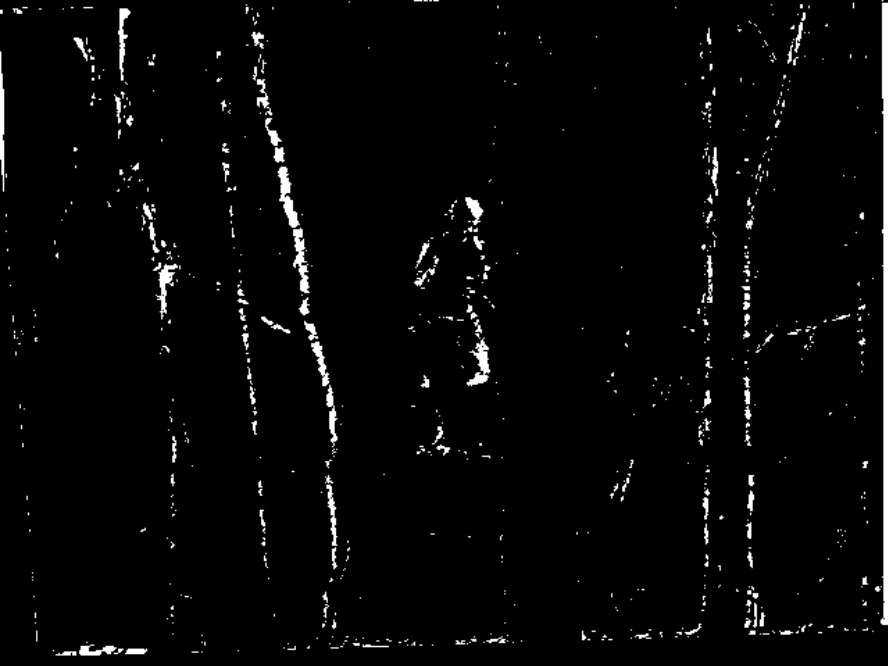}
		\subcaption{Motion parallax (*)}
	\end{minipage}
	\begin{minipage}{\textwidth}
		\centering
		\includegraphics[width=0.2\textwidth]{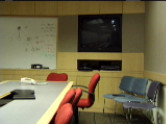}
		\includegraphics[width=0.2\textwidth]{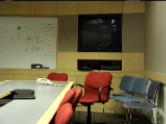}
		\includegraphics[width=0.2\textwidth]{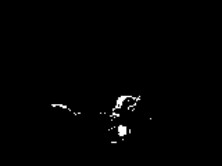}
		\subcaption{Moved background object}
	\end{minipage}
	\begin{minipage}{\textwidth}
		\centering
		\includegraphics[width=0.2\textwidth]{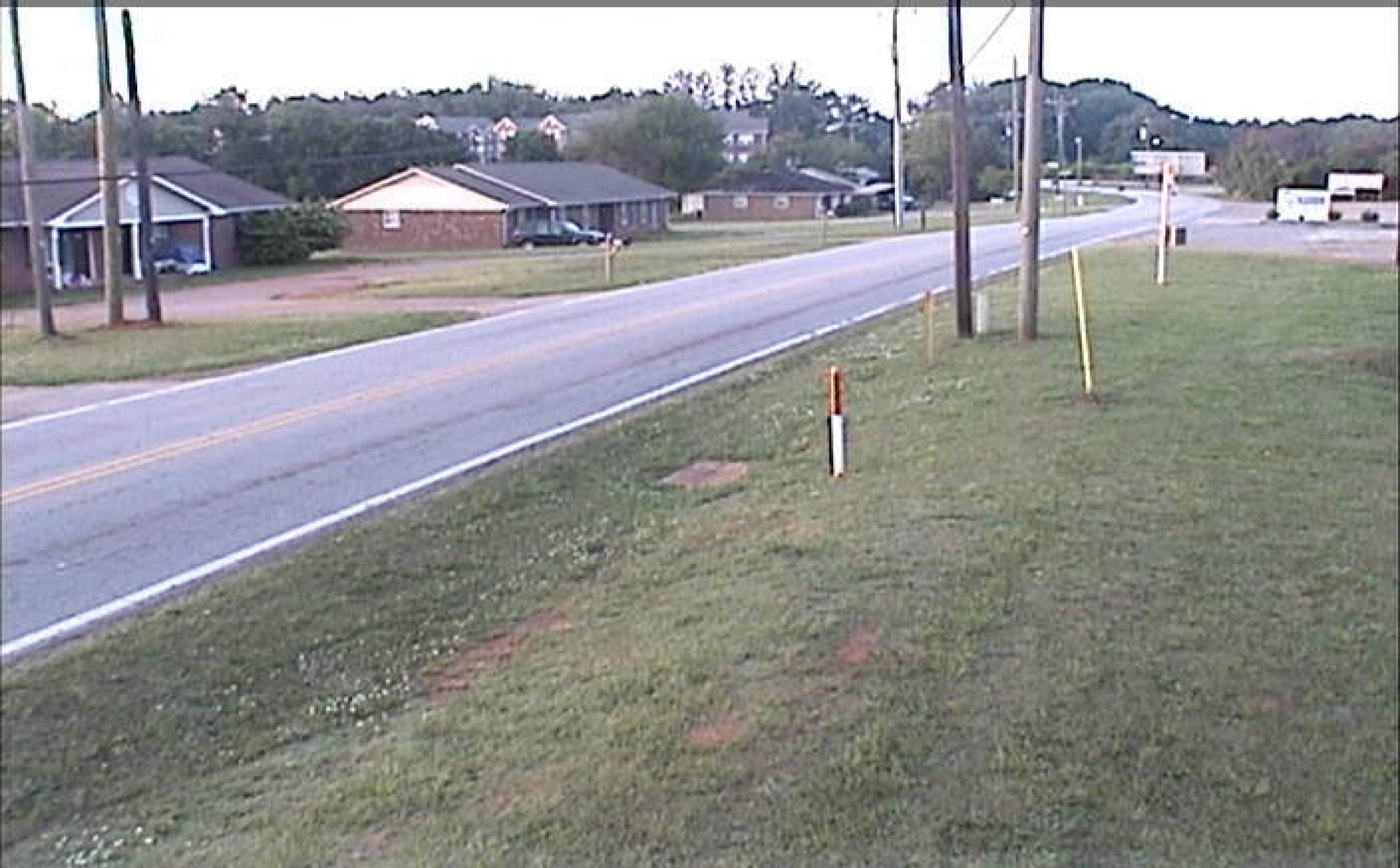}
		\includegraphics[width=0.2\textwidth]{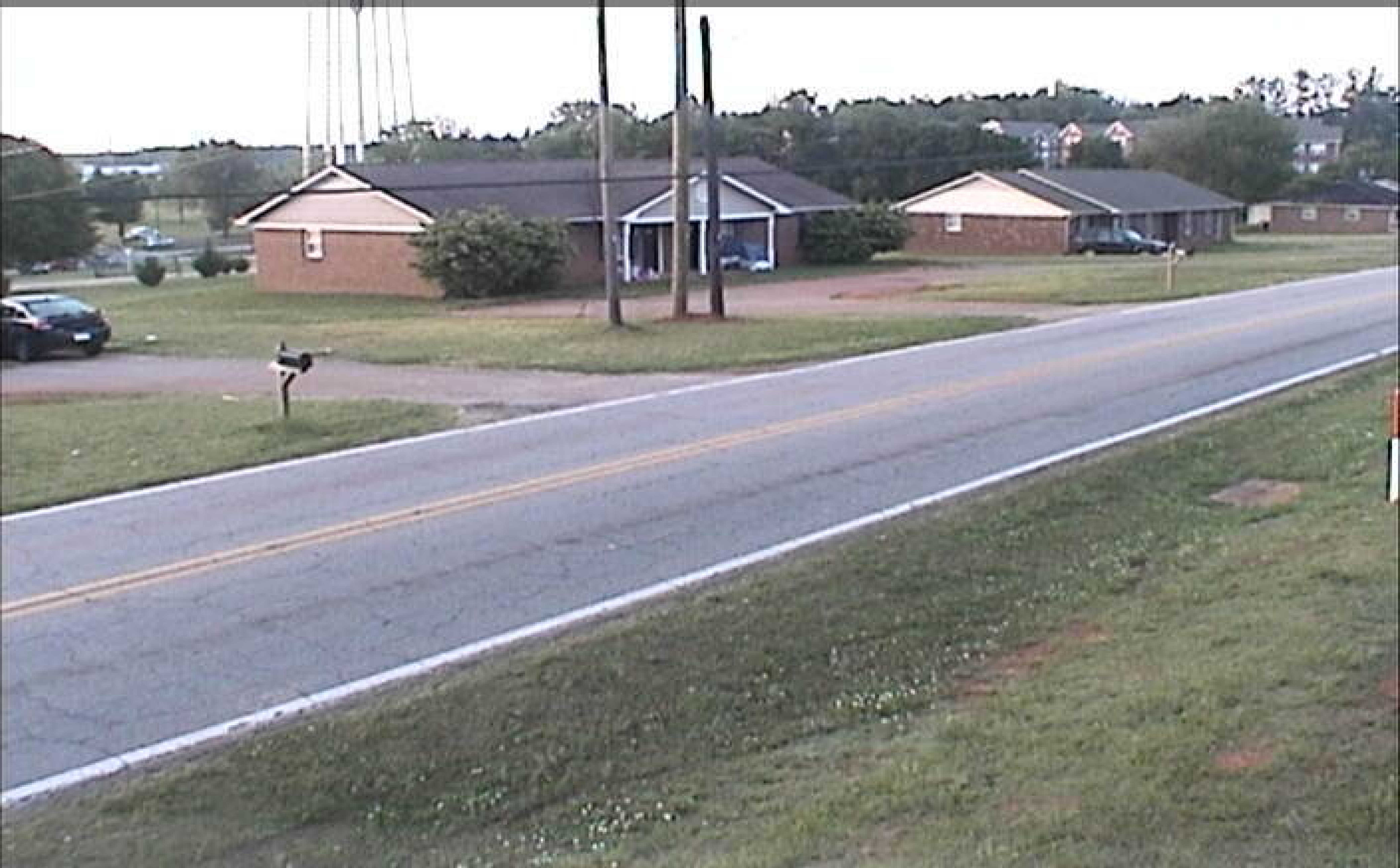}
		\includegraphics[width=0.2\textwidth]{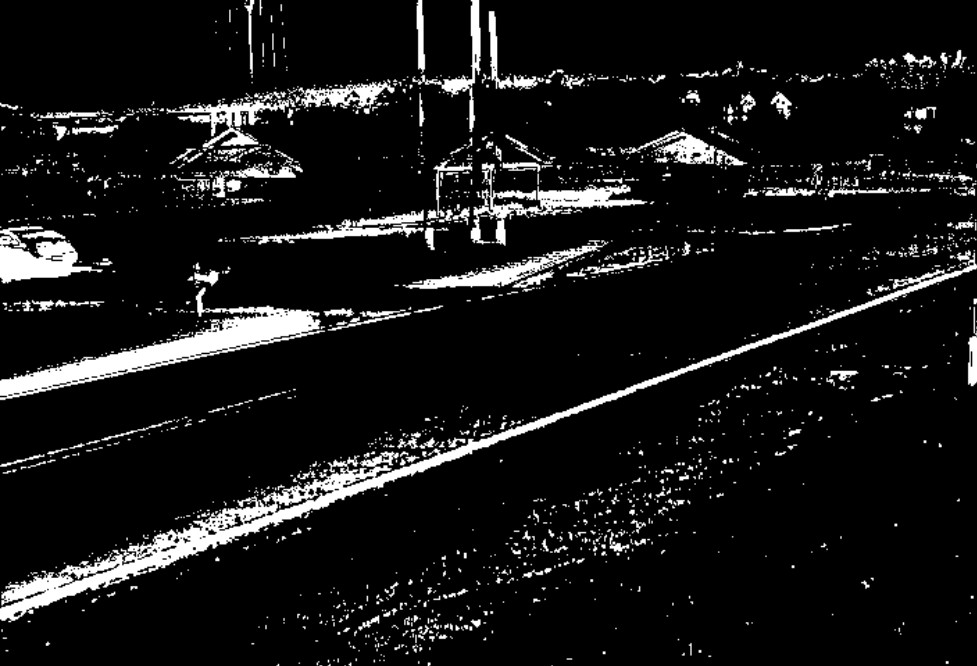}
		\subcaption{Moving camera}
	\end{minipage}
	\begin{minipage}{\textwidth}
		\centering
		\includegraphics[width=0.2\textwidth]{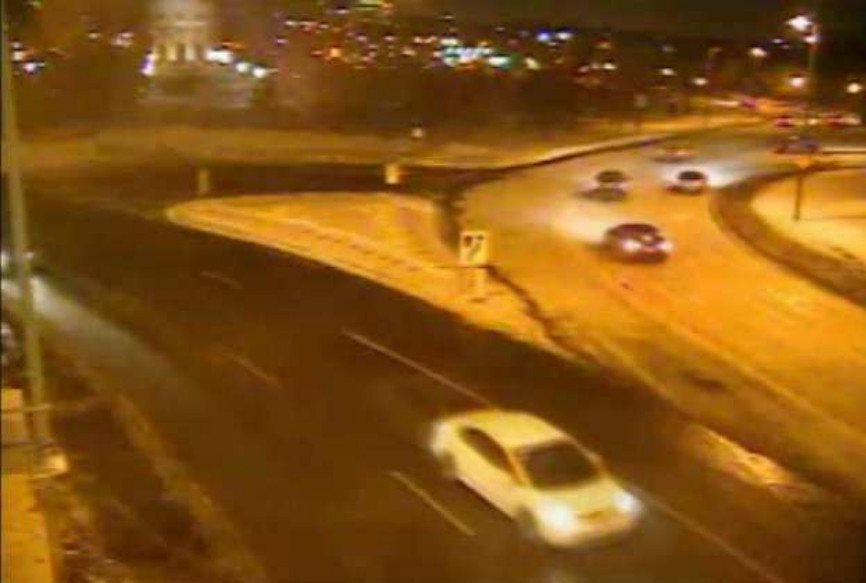}
		\includegraphics[width=0.2\textwidth]{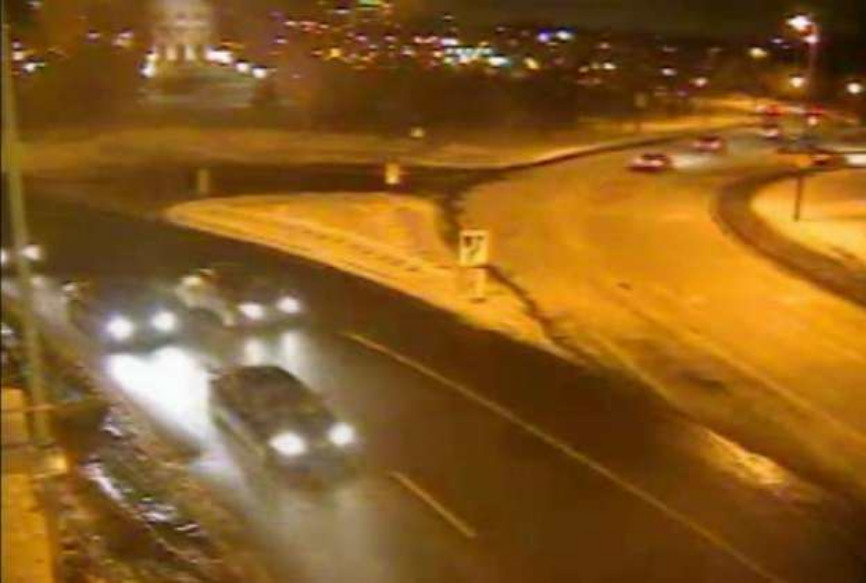}
		\includegraphics[width=0.2\textwidth]{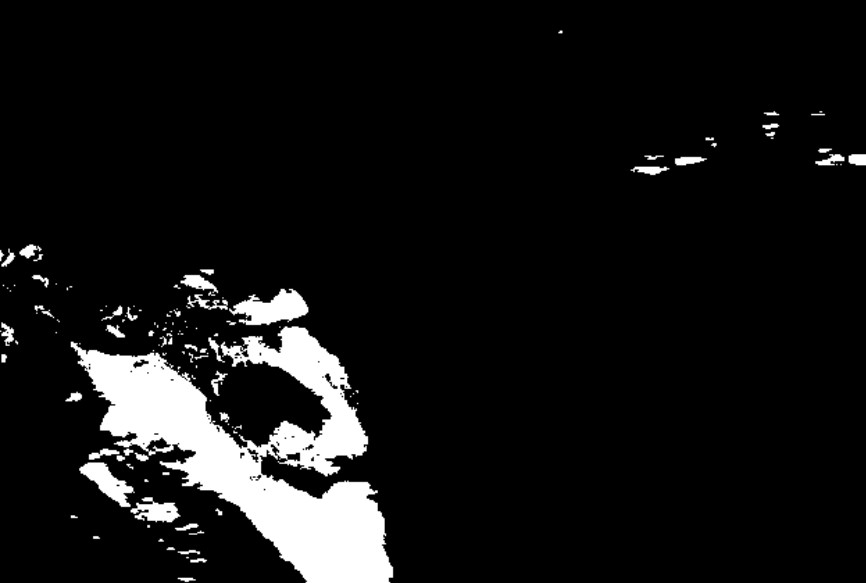}
		\subcaption{Night video}
	\end{minipage}
	\begin{minipage}{\textwidth}
		\centering
		\includegraphics[width=0.2\textwidth]{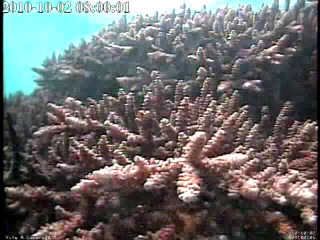}
		\includegraphics[width=0.2\textwidth]{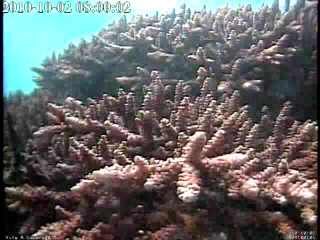}
		\includegraphics[width=0.2\textwidth]{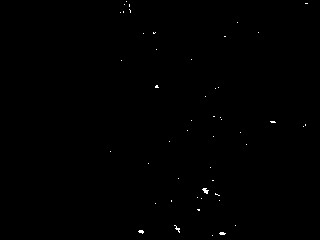}
		\subcaption{Noisy images}
	\end{minipage}
	\caption{Illustations of background subtraction challenges. Images come from the Wallflower (MovedObject sequence), the ChangeDetection.net (badminton, continuousPan, busyBoulvard sequences) and the ComplexBackground (Forest) dataset and the Fish4Knowledge (site NPP-3, camera 3, 10/02/2010 sequence) dataset. The last column is the result of Gaussian mixture-based background/foreground segmentation in the OpenCV library. (* To illustrate the Motion Parallax challenge, frames are register to the first one with a homography estimated by RANSAC on feature points.)}
	\label{fig:bs_challenges_2}
\end{figure*}

\begin{figure*}
	\begin{minipage}{\textwidth}
		\centering
		\includegraphics[width=0.2\textwidth]{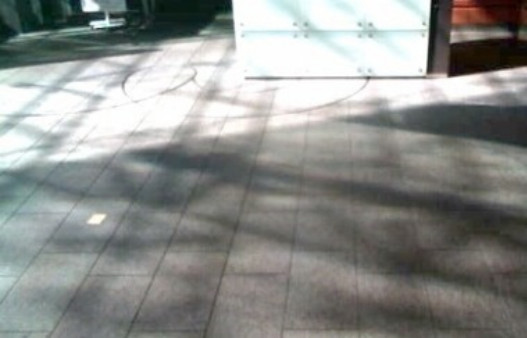}
		\includegraphics[width=0.2\textwidth]{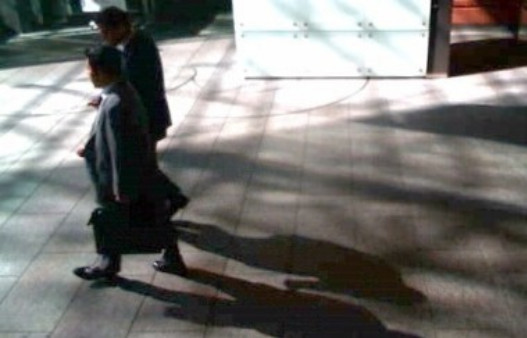}
		\includegraphics[width=0.2\textwidth]{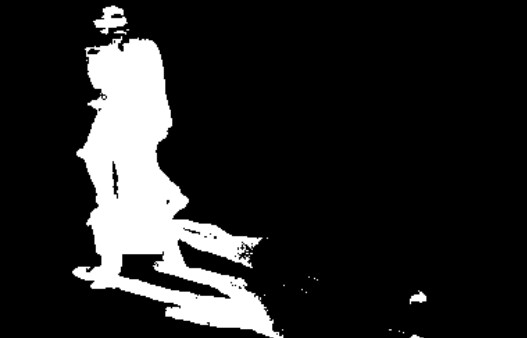}
		\subcaption{Shadows}
	\end{minipage}
	\begin{minipage}{\textwidth}
		\centering
		\includegraphics[width=0.2\textwidth]{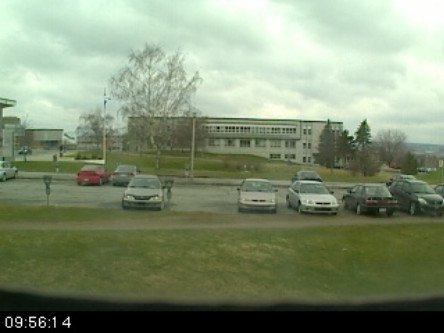}
		\includegraphics[width=0.2\textwidth]{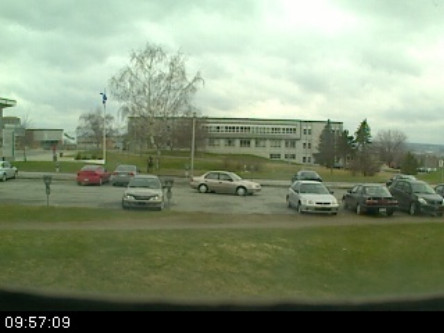}
		\includegraphics[width=0.2\textwidth]{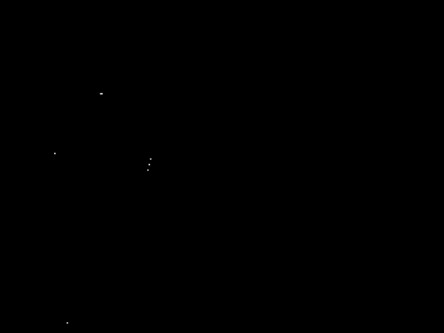}
		\subcaption{Sleeping foreground object}
	\end{minipage}
	\begin{minipage}{\textwidth}
		\centering
		\includegraphics[width=0.2\textwidth]{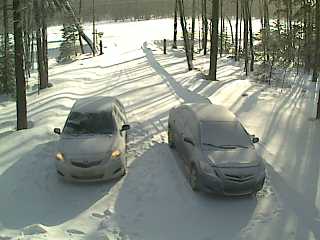}
		\includegraphics[width=0.2\textwidth]{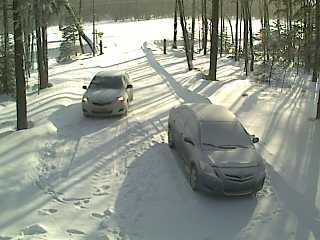}
		\includegraphics[width=0.2\textwidth]{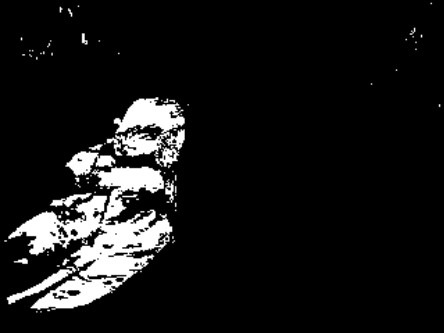}
		\subcaption{Waking foreground object}
	\end{minipage}
	\caption{Illustations of background subtraction challenges. Images come from the ChangeDetection.net (PeopleInShade, parking, winterDriveway sequences) dataset. The last column is the result of Gaussian mixture-based background/foreground segmentation in the OpenCV library.}
	\label{fig:bs_challenges_3}
\end{figure*}

Background subtraction is still an open issue with several scientific obstacles to overcome. In 1999, Toyama et al. \cite{Toyama1999} propose a list of 10 challenges about background maintenance for video surveillance systems. In this section we provide an extended list about the background subtraction challenges. Each challenge is illustrated by the figures~\ref{fig:bs_challenges_1}, \ref{fig:bs_challenges_2} and \ref{fig:bs_challenges_3}.
\begin{itemize}
\item \textbf{Bootstrapping} The training sequence doesn't contain only the background but also foreground objects.
\item \textbf{Camouflage} Foreground objects can have the same color than the background and become mixed up with it.
\item \textbf{Dynamic background} The background can contain some elements which are not completely static as water surface or waving trees. Even if there are not static, these elements are part of the background.
\item \textbf{Foreground aperture} The homogeneous part of a moving object cannot be detected and causes false negatives.
\item \textbf{Illumination changes} The difference of illumination between the current frame and the background model causes false detections. Illumination changes can be gradual (a cloud in front of the sun) or abrupt (light switch).
\item \textbf{Low frame rate} Background changes and illumination changes are not updated continuously with a low frame rate and these variations appear more abrupt.
\item \textbf{Motion blur} Images taken by the camera can be blurred by an abrupt camera motion or by camera jittering. 
\item \textbf{Motion parallax} 3D scenes with large depth variations present parallax in images taken by a moving camera. This parallax creates problem in background modeling and motion compensation.
\item \textbf{Moving camera} In a stationary camera, static objects appear static and moving objects appear moving. In the case of a moving camera, everything appears moving because of the camera displacement, also called the ego-motion. In these conditions it is more complicated to separate moving objects from the static ones. 
\item \textbf{Moved background object} Static objects can be move. These objects should not be considered as foreground.
\item \textbf{Night video} Images taken at night time present low brightness, low contrast and few color information.
\item \textbf{Noisy images} Noise in image depends on the quality of the camera components like sensors, lenses, resolutions.
\item \textbf{Shadows} Every objects create shadows by the interception of light rays. For a moving object, its shadow is moving but it must not be detected as foreground and it must not be integrated into the background model.
\item \textbf{Sleeping foreground object} When an object stops moving, it merges into the background.
\item \textbf{Waking foreground object} When an object starts to move a long time after the beginning of the video, the newly moving object and its old position in the background, called ghost, are detected as foreground.
\end{itemize}

Following these remarks, we can categorize the challenges by level of difficulties \cite{P0C0-Survey-200}. In addition, these challenges are less or more predominant depending on the real-applications \cite{P0C0-Survey-290}. For example in surveillance in natural environments like in maritime and aquatic environments, illumination changes and dynamic changes in the background are very challenging requiring more robust background methods than the top methods of CDnet 2014 as developed by Prasad et al. \cite{P0C0-A-29-2,P0C0-A-29-3,P0C0-A-29-3-1}. However, several authors provided tools to visualize and analyze the variations causes by theses challenges in the temporal history of the pixel \cite{P6C2-Tools-1,P6C2-Tools-10}.

	\newpage
%
\section{Static Cameras}
\label{sec:background_subtraction_with_a_stationary_camera}
There are three main categories of approaches to detect moving objects: consecutive frame difference, background subtraction, and optical flow. Consecutive frame difference methods \cite{P7C11-1,P7C11-10,P7C11-11} are very simple to implement but they are too sensitive to the challenges. Optical flow methods are more robust but are still too time consuming to reach real-time requirements. Background subtraction which is the most popular method to detect moving objects offers the best compromise between robustness and real-time requirements. In the literature, there exist a plenty of methods to detect moving objects by background subtraction and we let readers refer to books \cite{P0C0-Book-1,P0C0-Book-2} and surveys that cover this problematic for more details \cite{P0C0-Survey-13,P0C0-Survey-14,P0C0-Survey-24,P0C0-Survey-26,P0C0-Survey-34}. In this section, we describe the general process of background subtraction, survey the corresponding methods, and also investigate the current and unsolved challenges. This part is crucial to well understand extensions of background subtraction methods in the case of moving cameras.

\begin{figure}[!ht]
	\includegraphics[width=\columnwidth]{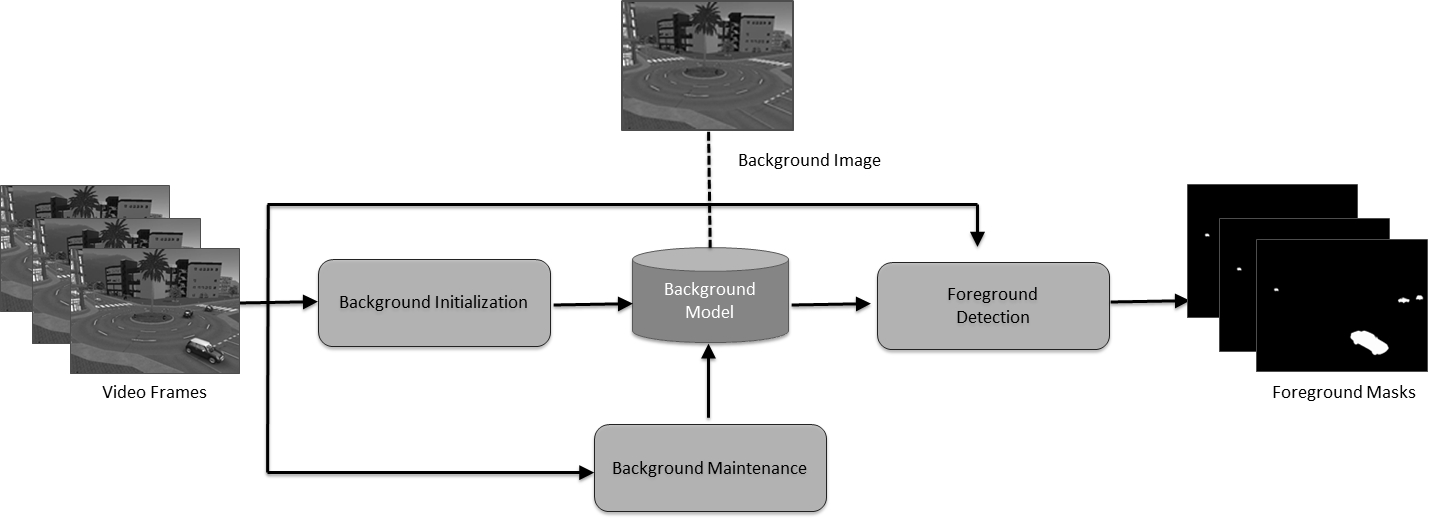}
	\caption{Background subtraction with a static camera, general scheme.}
	\label{fig:background_subtraction_static_camera}
\end{figure}

As defined in Section~\ref{sec:definitions}, from a static point of view only moving objects are moving. From this statement, background subtraction methods follow the general process (See Figure~\ref{fig:background_subtraction_static_camera}). Here, we describe the main process of each step.

\subsection{Background Modeling} 
The background model describes the model use to represent the background. A big variety of models coming from mathematical theories, machine learning and signal processing have been used for background modeling, including crisp models \cite{P1C1-Mean-1,P1C1-Median-300,P1C1-Histogram-27}, statistical models \cite{P1C2-KDE-2,P1C2-MOG-1000,P1C2-MOG-10,P1C2-MOG-636}, fuzzy models \cite{P5C1-FA-13,P2C2-10,P2C2-11}, Dempster-Schafer models \cite{P5C1-FA-70}, subspace learning models \cite{P1C4-1,P2C3-3,P2C3-1,P2C3-2,P2C3-100}, robust learning models \cite{P3C1-PCP-1,P3C1-PCP-917,P3C1-PCP-940,P3C1-PCP-941}, neural networks models \cite{P1C5-417,P1C5-415,P1C5-1} and filter based models \cite{P1C7-300,P1C7-200,P1C7-106,P1C7-1}.

\subsubsection{Mathematical models}
Based from mathematical theories, the simplest way to model a background is to compute the temporal average \cite{P1C1-Mean-1}, the temporal median \cite{P1C1-Median-300} or the histogram over time \cite{P1C1-Histogram-27}. These methods were widely used in traffic surveillance in 1990s owing to their simplicity but are not robust to the challenges faced in video surveillance such as camera jitter, changes in illumination, and dynamic backgrounds. To consider the imprecision, uncertainty and incompleteness in the observed data (i.e. video), statistical models began being introduced in 1999 such as single Gaussian \cite{P1C2-SG-1}, Mixture of Gaussians (MOG) \cite{P1C2-MOG-1000,P1C2-MOG-10} and Kernel Density Estimation \cite{P1C2-KDE-2,P1C2-KDE-30}. These methods based on a Gaussian distribution model proved to be more robust to dynamic backgrounds \cite{P1C2-MOG-760,P1C2-MOG-780}. More advanced statistical models were after developed in the literature and can be classified into those based on another distribution that alleviate the strict Gaussian constraint (i.e. general Gaussian distribution \cite{P2C1-14}, Student's t-distribution \cite{P2C1-1,P2C1-2}, Dirichlet distribution \cite{P2C1-11,P2C1-12}, Poisson distribution \cite{P2C1-20,P2C1-30}), those based on co-occurrence \cite{P2C0-1400,P2C1-260,P2C1-261} and confidence \cite{P2C0-6000,P2C0-6001}, free-distribution models \cite{P2C1-200,P2C0-40,P2C0-50}, and regression models \cite{P2C0-5100,P2C0-5101}. These approaches have improved the robustness to various challenges over time. The most accomplished methods in this statistical category are ViBe \cite{P2C1-200}, PAWCS \cite{P2C0-50} and SubSENSE \cite{P2C0-40}. Another theory that allows the handling of imprecision, uncertainty, and incompleteness is based on the fuzzy concept. In 2006-2008, several authors employed concepts like Type-2 fuzzy sets \cite{P2C2-10,P2C2-12,P2C2-15}, Sugeno integral \cite{P5C1-FA-10,P5C1-FA-11} and Choquet integral \cite{P5C1-FA-12,P5C1-FA-13,P5C1-FA-40}. These fuzzy models show robustness in the presence of dynamic backgrounds \cite{P2C2-12}. Dempster-Schafer concepts were also be employed in foreground detection \cite{P5C1-FA-70}.

\subsubsection{Machine learning models}
Based on machine learning, background modeling has been investigated by representation learning (also called subspace learning), support vector machines, and neural networks modeling (conventional and deep neural networks). 
\begin{itemize}
\item \textbf{Representation learning:} In 1999, reconstructive subspace learning models like Principal Component Analysis (PCA) \cite{P1C4-1} has been introduced to learn the background in an unsupervised manner. Subspace learning models handle illumination changes more robustly than statistical models \cite{P0C0-Survey-23}. In further approaches, discriminative \cite{P2C3-3,P2C3-1,P2C3-2} and mixed \cite{P2C3-100} subspace learning models have been used to increase the performance for foreground detection. However, each of these regular subspace methods presents a high sensitivity to noise, outliers, and missing data. To address these limitations, since 2009, a robust PCA through decomposition into low-rank plus sparse matrices \cite{P3C1-PCP-1,P3C1-PCP-917,P3C1-PCP-940,P3C1-PCP-941} has been widely used in the field. These methods are not only robust to changes in illumination but also to dynamic backgrounds \cite{P3C1-PCP-942,P3C1-PCP-9420,P3C1-RMC-90,P3C1-RMC-91}. However, they require batch algorithms, making them impractical for real-time applications. To address this limitation, dynamic robust PCA as well as robust subspace tracking \cite{P3C1-PCP-1030-1,P3C1-PCP-1030,P3C1-PCP-1096-4} have been designed to achieve a real-time performance of RPCA-based methods. The most accomplished methods in this subspace learning category are GRASTA \cite{P3C2-RST-2}, incPCP \cite{P3C1-PCP-1063}, ReProCS \cite{P3C1-PCP-1018} and MEROP \cite{P3C1-PCP-1027-1}. However, tensor RPCA based methods \cite{P3C5-TRPCA-IT-3,P3C5-TRPCA-IT-4,P3C5-TRPCA-IT-100,P3C5-TRPCA-IT-110} allow to take into account spatial and temporal constraints making them more robust against noise.
\item \textbf{Neural networks modeling:} In 1996, Schofield et al. \cite{P1C5-1} were the first to use neural networks for background modeling and foreground detection through the application of a Random Access Memory (RAM) neural network. However, a RAM-NN requires the images to represent the background of the scene correctly, and there is no background maintenance stage because once a RAM-NN is trained with a single pass of background images, it is impossible to modify this information. In 2005, Tavakkoli \cite{P1C5-20} proposed a neural network approach under the concept of novelty detector. During the training step, the background is divided in blocks. Each block is associated to a Radial Basis Function Neural Network (RBF-NN). Thus, each RBF-NN is trained with samples of the background corresponding to its associated block. The decision of using RBF-NN is because it works like a detector and not a discriminant, generating a close boundary for the known class. RBF-NN methods is able to address dynamic object detection as a single class problem, and to learn the dynamic background. However, it requires a huge amount of samples to represent general background scenarios. In 2008, Maddalena and Petrosino \cite{P1C5-400,P1C5-401,P1C5-402,P1C5-404} proposed a method called Self Organizing Background Subtraction (SOBS) based on a 2D self-organizing neural network architecture preserving pixel spatial relations. The method is considered as nonparametric, multi-modal, recursive and pixel-based. The background is automatically modeled through the neurons weights of the network. Each pixel is represented by a neural map with $n \times n$ weight vectors. The weights vectors of the neurons are initialized with the corresponding color pixel values using the HSV color space. Once the model is initialized, each new pixel information from a new video frame is compared to its current model to determine if the pixel corresponds to the background or to the foreground. In further works, SOBS was improved in several variants such as Multivalued SOBS \cite{P1C5-405}, SOBS-CF \cite{P1C5-407}, SC-SOBS \cite{P1C5-408}, 3dSOBS+ \cite{P1C5-410}, Simplified SOM \cite{P1C5-412}, Neural-Fuzzy SOM \cite{P1C5-414} and MILSOBS \cite{P1C5-421}) which allow this method to be in the leading methods on the CDnet 2012 dataset \cite{P6C2-Dataset-1000} during a long time. SOBS show also interesting performance for stopped object detection \cite{P1C5-403,P1C5-406,P1C5-409}. But, one of the main disadvantages of SOBS based methods is the need to manual adjust at least four parameters. 
\item \textbf{Deep Neural networks modeling:} Since 2016, DNNs have also been successfully applied to background generation \cite{P1C5-1900,P1C5-2130,P1C5-1910,P1C5-2001,P1C5-2000}, background subtraction \cite{P1C5-2140,P1C5-2120,P1C5-2100,P1C5-2160,P1C5-2150,P1C5-2090,P1C5-2091,P1C5-2092}, foreground detection enhancement \cite{P1C5-2175}, ground-truth generation \cite{P1C5-2110}, and the learning of deep spatial features \cite{P1C5-2173,P1C5-2171,P1C5-2200,P1C5-2210,P1C5-2010}. More practically, Restricted Boltzman Machines (RBMs) were first employed by Guo and Qi \cite{P1C5-1900} and Xu et al. \cite{P1C5-1910} for background generation to further achieve moving object detection through background subtraction. In a similar manner, Xu et al. \cite{P1C5-2001,P1C5-2000} used deep auto-encoder networks to achieve the same task whereas Qu et al. \cite{P1C5-2130} used context-encoder for background initialization. As another approach, Convolutional Neural Networks (CNNs) has also been employed to background subtraction by Braham and Droogenbroeck \cite{P1C5-2100}, Bautista et al. \cite{P1C5-2120} and Cinelli \cite{P1C5-2160}. Other authors have employed improved CNNs such as cascaded CNNs \cite{P1C5-2110}, deep CNNs \cite{P1C5-2140}, structured CNNs \cite{P1C5-2150} and two stage CNNs \cite{P1C5-2161}. Through another approach, Zhang et al. \cite{P1C5-2010} used a Stacked Denoising Auto-Encoder (SDAE) to learn robust spatial features and modeled the background with density analysis, whereas Shafiee et al. \cite{P1C5-2200} employed Neural Reponse Mixture (NeREM) to learn deep features used in the Mixture of Gaussians (MOG) model \cite{P1C2-MOG-10}. In 2019, Chan \cite{P1C5-2189-4} proposed a deep learning-based scene-awareness approach for change detection in video sequences thus applying the suitable background subtraction algorithm for the corresponding type of challenges.
\end{itemize}

\subsubsection{Signal processing models}
Based on signal processing, these models considered temporal history of a pixel as 1-D dimensional signal. Thus, several signal processing methods can be used: 1) signal estimation models (i.e. filters), 2) transform domain functions, and 3) sparse signal recovery models (i.e. compressive sensing). 
\begin{itemize}
\item \textbf{Estimation filter:} In 1990, Karmann et al. \cite{P1C7-100} proposed a background estimation algorithm based on the Kalman filter. Any pixel that deviates significantly from its predicted value is declared foreground. Numerous variants were proposed to improve this approach in the presence of illumination changes and dynamic backgrounds \cite{P1C7-105,P1C7-106,P1C7-110}. In 1999, Toyama et al. \cite{P1C7-1} proposed in their algorithm called Wallflower a pixel-level algorithm which makes probabilistic predictions about what background pixel values are, expected in the next live image using a one-step Wiener prediction filter. Chang et al. \cite{P1C7-300,P1C7-301} used a Chebychev filter to model the background. All these filters approaches reveal good performance in the presence of slow illumination change but less when the scenes present complex dynamic backgrounds.
\item \textbf{Transform domain models:} In 2005, Wren and Porikli \cite{P5C1-TDF-1} estimated the background model that captures spectral signatures of multi-modal backgrounds using Fast Fourier Transform (FFT) features through a method called Waviz. Here, FFT features are then used to detect changes in the scene that are inconsistent over time. In 2005, Porikli and Wren \cite{P5C1-TDF-2} developed an algorithm called Wave-Back that generated a representation of the background using the frequency decompositions of pixel history. The Discrete Cosine Transform (DCT) coefficients are used as features are computed for the background and the current images. Then, the coefficients of the current image are compared to the background coefficients to obtain a distance map for the image. Then, the distance maps are fused in the same temporal window of the DCT to improve the robustness against noise. Finally, the distance maps are thresholded to achieve foreground detection. This algorithm is efficient in the presence of waving trees. 
\item \textbf{Sparse signal recovery models:} In 2008, Cevher et al. \cite{P3C4-CS-1} were the first authors who employed a compressive sensing approach for background subtraction. Instead of learning the full background, Cevher et al. \cite{P3C4-CS-1} learned and adapted a low dimensional compressed representation of it which is sufficient to capture changes. Then, moving objects are estimated directly using the compressive samples without any auxiliary image reconstruction. But, to obtain simultaneously appearance recovery of the objects using compressive measurements, it needs to reconstruct one auxiliary image. To alleviate this constraint, numerous improvements were proposed in the literature \cite{P3C4-CS-5,P3C4-CS-30,P3C4-CS-31,P3C4-CS-70,P3C4-CS-103} and particular good performance is obtained by Bayesian compressive sensing approaches \cite{P3C4-CS-110,P3C4-CS-111,P3C4-CS-112,P3C4-CS-113}.
\end{itemize}

\subsection{Background initialization}
This step consists in computing the first background image and it is also called \textit{background generation}, \textit{background extraction} and \textit{background reconstruction}. The background model is initialized with a set of images taken before the moving objects detection process. Several kind of models could be used to initialize the background and they are classified as methods based on temporal statistics \cite{P4C1-1,P4C1-4,P4C1-23}, methods based on sub-sequences of stable intensity \cite{P4C1-2,P4C1-7,P4C1-3,P4C1-8,BI-5,BI-5-1,BI-5-2}, methods based on missing data reconstruction problem \cite{P3C1-PCP-3070,P3C1-PCP-3070-1}, methods based on iterative model completion \cite{P1C3-1}, methods based on conventional neural networks \cite{P1C5-408,P1C5-1603}, and methods based on optimal labeling \cite{P4C1-100}. The most accomplished methods applied to the SBMnet dataset \cite{P0C0-Survey-33} are Motion-assisted Spatio-temporal Clustering of Low-rank (MSCL) designed by Javed et al. \cite{P3C1-PCP-9420}, and LaBGen and its variants developed by Laugraud et al. \cite{BI-5,BI-5-1,BI-5-2}. For more details, the reader can refer to comprehensive surveys of Maddelena and Petrosino \cite{P0C0-Survey-30,P0C0-Survey-31,P0C0-Survey-32,P0C0-Survey-33}.

\subsection{Updating background model}
In order to overcome background changes (illumination changes, dynamic background, and so on), the background model is updated with information provided by the current frame taken by the camera. The update rules depend on the model chosen but they generally try to employ old data with the new one according to a learning rate. The choose of the learning rate allow to integrate more or less rapidly the changes to the background. The maintenance of the background model is a critical step since some parts of the foreground could be integrated in the background and create false-alarms. However, the background maintenance process requires an incremental on-line algorithm, since new data is streamed and so dynamically provided. The key issues of this step are the following ones:
\begin{itemize}
\item\textbf{Maintenance schemes:} In the literature, three maintenance schemes are present: the blind, the selective, and the fuzzy adaptive schemes \cite{P4C3-F-1}. The blind background maintenance updates all the pixels with the same rules which is usually an IIR filter. The main disadvantage of this scheme is that the value of pixels classified as foreground are used in the computation of the new background and so polluted the background image. To solve this problem, some authors used a selective maintenance scheme that consists of updating the new background image with different learning rate depending on the previous classification of a pixel into foreground or background. Here, the idea is to adapt very quickly a pixel classified as background and very slowly a pixel classified as foreground. But the problem is that erroneous classification may result in a permanent incorrect background model.
This problem can be addressed by a fuzzy adaptive scheme which takes into account the uncertainty of the classification. This can be achieved by graduating the update rule using the result of the foreground detection such as in El Baf et al. \cite{P4C3-F-1}. 
\item\textbf{Learning rate:} The learning rate determines the speed of the adaptation to the scene changes. It can be fixed, or dynamically adjusted by a statistical, or a fuzzy method. In the first case, the learning rate is fixed as the same value for all the sequence. Then, it is determined carefully such as in \cite{P1C2-MOG-78} or can be automatically selected by an optimization algorithm \cite{P1C2-MOG-79}. However, it can take one value for the learning step and one for the maintenance step \cite{P1C2-MOG-90}. Additionally, the rate may change over time following a tracking feedback strategy \cite{P1C2-MOG-110}. For the statistical case, Lee \cite{P1C2-MOG-100} used different learning rates for each Gaussian in the MOG model. The convergence speed and approximation results are significantly improved. For the fuzzy case \textbf{(3)}, Sigari et al. \cite{P4C2-F-1,P4C2-F-2} computed an adaptive learning rate at each pixel with respect to the fuzzy membership value obtained for each pixel during the fuzzy foreground detection. In another way, Maddalena and Petrosino \cite{P1C5-405,P1C5-407} improved the adaptivity by introducing spatial coherence information.
\item \textbf{Maintenance mechanisms:} The learning rate determines the speed of adaptation to illumination changes but also the time a background change requires until it is incorporated into the model as well as the time a static foreground object can survive before being included in the model. So, the learning rate deals with different challenges which have different temporal characteristics. To decouple the adaptation mechanism and the incorporation mechanism, some authors \cite{P1C2-MOG-70}\cite{P1C2-MOG-73} used a set of counters which represents the number of times a pixel is classified as a foreground pixel. When this number is larger than a threshold, the pixel is considered as background. This gives a time limit on how long a pixel can be considered as a static foreground pixel. 
\item\textbf{Frequency of the update:} The aim is to update only when it is needed. The maintenance may be done every frame but in absence of any significant changes, pixels are not required to be updated at every frame. For example, Porikli \cite{P1C2-MOG-102} proposed adapting the time period of the maintenance mechanism with respect to an illumination score change. The idea is that no maintenance is needed if no illumination change is detected and a quick maintenance is necessary otherwise. In the same idea, Magee \cite{P1C2-MOG-195} used a variable adaptation frame rate following the activity of the pixel, which improves temporal history storage for slow changing pixels while running at high adaption rates for less stable pixels. 
\end{itemize}

\subsection{Foreground detection}
As the name of the technique suggests it, the foreground is detected by subtracting the background to the current frame. A too high difference, determined by a threshold, points the foreground out. The output is a binary image so-called a mask for which each pixel is classified as background or foreground. Thus, this task is a classification one, that can be achieved by crisp, statistical or fuzzy classification tools. For this, the different steps have to be achieved:
\begin{itemize}
\item \textbf{Pre-processing:} The pre-processing step avoids the detection of unimportant changes due to the motion of the camera or the illumination changes. This step may involve geometric and intensity adjustments \cite{P4C2-CD-1100}. As the scenes are usually rigid in nature and the camera jitter is small, geometric adjustments can often be performed using low-dimensional spatial transformations such as similarity, affine, or projective transformations \cite{P4C2-CD-1100}. On the other hand, there are several ways to achieve intensity adjustments. This can be done with intensity normalization \cite{P4C2-CD-1100}. The pixel intensity values in the current image are then normalized to have the same mean and variance as those in the background image. Another way consists in using a homomorphic filter based which is based on the shading model. This approach permits to separate the illumination and the reflectance. As only the reflectance component contains information about the objects in the scene, illumination-invariant foreground detection \cite{P5C1-CDFeatureI-303,P5C1-CDFeatureI-304,P5C1-CDFeatureI-310} can hence be performed by first filtering out the illumination component from the image.
\item \textbf{Test:} The test which allows to classify pixels of the current image as background or foreground is usually the difference between the background image and the current image. This difference is then thresholded. Another way to compare two images are the significance and hypothesis tests. The decision rule is then cast as a statistical hypothesis test. The decision as to whether or not a change has occurred at a given pixel corresponds to choosing one of two competing hypotheses: the null hypothesis $H_0$ or the alternative hypothesis $H_1$, corresponding to no-change and change decisions, respectively. Several significance tests can be found in the literature \cite{P5C1-CDFeatureI-1,P5C1-CDFeatureI-3,P5C1-CDFeatureI-300,P5C1-CDFeatureI-301,P5C1-CDFeatureI-302,P5C1-CDFeatureI-400,P5C1-CDFeatureI-401,P5C1-CDFeatureI-402}.
\item \textbf{Threshold:} In literature, there are several types of threshold schemes. First, the threshold can be fixed and the same for all the pixels and the sequence. This scheme is simple but not optimal. Indeed, pixels present different activities and it needs an adaptive threshold. This can be done by computing the threshold via the local temporal standard deviation of intensity between the background and the current images, and by updating it using an infinite impulse response (IIR) filter such as in Collins et al. \cite{P4C2-CFD-1}. An adaptive threshold can be statistically obtained also from the variance of the pixel such as in Wren et al. \cite{P1C2-SG-1}. Another way to adaptively threshold is to use fuzzy thresholds such as in the studies of Chacon-Muguia and Gonzalez-Duarte \cite{P1C5-413}.
\item \textbf{Post-processing:} The idea here is to enhance the consistency of the foreground mask. This can be done firstly by deleting isolated pixels with classical or statistical morphological operators \cite{P4C2-CD-1101}. Another way is to use fuzzy concepts such as fuzzy inference between the previous and the current foreground masks \cite{P4C2-PP-1}.
\end{itemize}

\bigskip

Moreover, foreground detection is a particular case of change detection when \textbf{(1)} one image is the background and the other one is the current image, and \textbf{(2)} the changes concern moving objects. So, all the techniques developed for change detection can be used in foreground detection. A survey concerning change detection can be found in \cite{P4C2-CD-1100,P4C2-CD-1102}. 

\subsubsection{Solved and Unsolved Challenges}
For fair evaluation and comparison on videos presenting challenges described in CDnet 2014 dataset \cite{P6C2-Dataset-1001} which was developed as part of Change Detection Workshop challenge (CDW 2014). This dataset includes all the videos from the CDnet 2012 dataset \cite{P6C2-Dataset-1000} plus 22 additional camera-captured videos providing 5 different categories that incorporate challenges that were not addressed in the 2012 dataset. The categories are as follows: baseline, dynamic backgrounds, camera jitter, shadows, intermittent object motion, thermal, challenging Weather, low frame-rate, night videos, PTZ and turbulence. In 2015, Jodoin \cite{P6C2-Dataset-1025} did the following remarks regarding the solved and unsolved challenges by using the experimental results available at CDnet 2014:
\begin{itemize}
\item Conventional background subtraction methods can efficiently deal with challenges met in "baseline" and "bad weather" sequences. 
\item The "Dynamic backgrounds", "thermal video" and "camera jitter" categories are a reachable challenge for top-performing background subtraction.
\item The "Night videos", "low frame-rate", and "PTZ" video sequences represent significant challenges.
\end{itemize}
However, Bouwmans et al. \cite{P0C0-Survey-27} analyzed the progression made over 20 years from the MOG model \cite{P1C2-MOG-10} designed in 1999 up to the recent deep neural networks models developed in 2019. To do so, Bouwmans et al. \cite{P0C0-Survey-27} computed different key increases in the F-measure score in terms of percentage by considering the gap between MOG \cite{P1C2-MOG-10} and the best conventional neural network (SC-SOBS \cite{P1C5-408}), the gap between SC-SOBS \cite{P1C5-408} and the best non-parametric multi-cues methods (SubSENSE \cite{P2C0-40}), the gap between SuBSENSE \cite{P2C0-40} and Cascaded CNNs \cite{P1C5-2110}, the gap between SuBSENSE \cite{P2C0-40} and the best DNNs based method (FgSegNet-V2 \cite{P1C5-2168}), and the gap between FgSegNet-V2 \cite{P1C5-2168} and the ideal method (F-Measure$=1$ in each category). The big gap has been obtained by DNNs methods against SuBSENSE with $24.31\%$ and $32.92\%$ using Cascaded CNN and FgSegNet-V2, respectively. The gap of $1.55\%$ that remains between FgSegNet-V2 and the ideal method is less than the gap of $6.93\%$ between Cascaded CNN and FgSegNet-V2. Nevertheless, it is important to note that the large gap provided by cascaded CNN and FgSegNet-V2 is mainly due to their supervised aspect, and a required drawback of training using labeling data. However, when labeling data are unavailable, efforts should be concentrated on unsupervised GANs as well as unsupervised methods based on semantic background subtraction \cite{10,11}, and robust subspace tracking \cite{P3C1-PCP-1027-1,P3C1-PCP-1096-4,P3C1-PCP-1064,P3C1-PCP-1063,P3C1-PCP-1030-1,P3C1-PCP-1030} that are still of interest in the field of background subtraction. 
Furthermore, deep learning approaches detect the changes in images with static backgrounds successfully but are more sensitive in the case of dynamic backgrounds and camera jitter, although they do provide a better performance than conventional approaches \cite{P1C5-2189-2}. In addition, several authors avoid experiments on the "IOM" and the "PTZ" categories. In addition, when the F-Measure score is provided for these categories, the score is not very high. Thus, it seems that the current deep neural networks tested face problems in theses cases perhaps because they have difficulties in how to learn the duration of sleeping moving objects and how to handle changes from moving cameras. 
However, even if background subtraction models designed for static cameras progress for camera jitter and PTZ cameras as with several RPCA models \cite{P3C1-PCP-1064,P3C1-PCP-1064-1,P3C1-PCP-1064-2,P3C2-RST-100,P3C2-RST-101,P3C2-1} and deep learning models \cite{P1C5-2168,P1C5-21680,P1C5-21681}, they can only handle small jitter movements or translation and rotation movements. Thus, detection of moving objects with moving cameras required more dedicated strategies and models that we reviewed in this survey.

%

\section{Moving Cameras}
\label{sec:methods_classification}

The background subtraction that we have just presented here is designed for static camera cannot be applied directly to moving camera since the background is no longer static in the images. Most of the methods that we present in this paper are adaptions or inspirations of the idea of background subtraction to a moving camera.

\begin{figure}[!ht]
	\centering
	\includegraphics[width=\columnwidth]{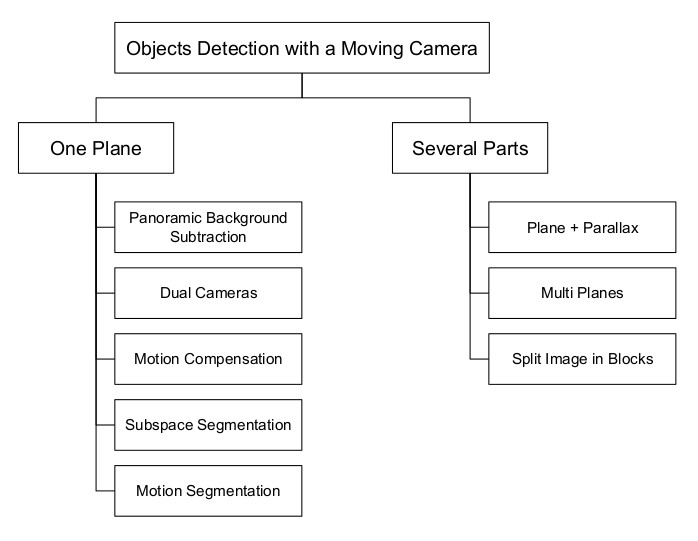}
	\caption{The taxonomy adopted in this survey.}
	\label{fig:taxonomy}
\end{figure}

In the case of a moving camera, the result of foreground detection depends on the background representation. We choose to categorize the methods of moving objects detection with a moving camera by the type of background representation chosen to solve the problem. The figure~\ref{fig:taxonomy} presents the taxonomy adopted in this survey.

%

\subsection{One plane}
The methods presented in this section represent the background as one plane in the presence of flat scenes. The methods are grouped together according to five different approaches.

%

\subsubsection{Panoramic background subtraction}
\label{sec:panoramic_background_subtraction}
\begin{figure}[!ht]
	\centering
	\includegraphics[width=\columnwidth]{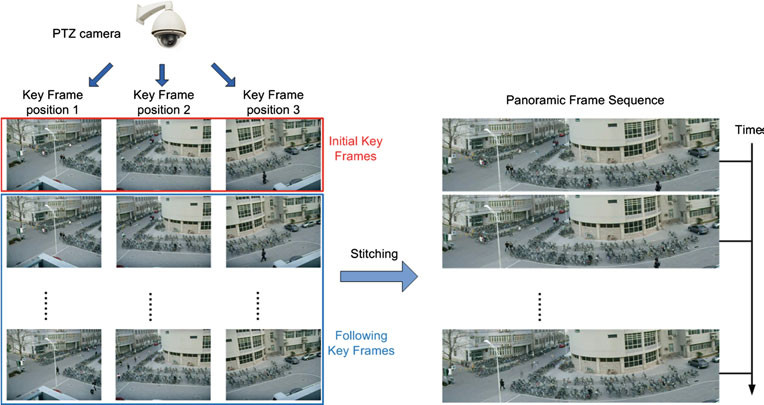}
	\caption{An example of a technique to construct a panoramic background model.\\ \textit{Source}: Images from Xue et al. \cite{Xue2013}.}
	\label{fig:panoramic_background_model_construction}
\end{figure}

The images captured by a moving camera can be stitch together to form a bigger image so-called a panorama or a mosaic as shown in the figure~\ref{fig:panoramic_background_model_construction} . This panorama can be used to model the background and detect moving objects as for a static camera.

The construction of a panoramic view is a key step that needs high accuracy \cite{Irani1996, Faugeras2001, Brown2003, Brown2007}. There are three techniques to align the images to construct the mosaic:
\begin{itemize}
\item \textbf{Frame to frame}: alignment parameters are computed for each pair of successive frames for the entire sequence. All the frames are then aligned to a fixed coordinate system, given by the reference frame or a virtual coordinate system. The problem with this mosaic construction is that errors may accumulate during the alignment to the fixed coordinate system.
\item \textbf{Frame to mosaic}: since the mosaic is larger than a frame, large displacement has to be handle to align a frame to the mosaic. To manage it, the parameters between the previous frame and the mosaic are used as an initial estimation since they are closed to those between the new frame and the mosaic.
\item \textbf{Mosaic to frame}: contrary to the two previous alignment techniques, the mosaic is aligned to the new frame. There is no static coordinate system and the current image is maintained in its input coordinate systems.
\end{itemize}
The two first techniques, \emph{frame-to-frame} and \emph{frame-to-mosaic} are widely used in the construction of a mosaic for the moving object detection problem.

\bigskip

In order to warp images to form a mosaic, two motions are generally used: the affine or the perspective motion model \cite{Brown1992, Zitova2003}. The perspective transformations better fit the camera transformation but in some cases the affine motion model can be sufficient and it is also faster since there are only six parameters to estimate against eight for the perspective one. In both cases, a refinement step is generally performed to correct misalignment errors.

\bigskip

In 2000, Mittal et al. \cite{Mittal2000} construct the panorama by registering an image to the entire mosaic in order to limit cascading of registration errors with the frame to frame technique. The registration is performed by an affine transformation based on the Kanade-Lucas-Tomasi (KLT) feature tracker \cite{Shi1994} which is refined by using the Levenberg-Marquardt method. In an other work, Bartoli et al. \cite{Bartoli2002} combine the direct method and the feature based method to construct a panorama. The feature based method is used to obtain a first estimation of the panorama. Then a direct method refines each frame registration. To deal with real-time and accuracy, Bevilacqua et al. \cite{Bevilacqua2005} use a feature-based method to construct a panorama where the outlier features are filtered by a simple but efficient clustering method in order to estimate a projective transformation with only features that result to the camera ego-motion. The frame-to-frame alignment errors are fixed by a two-stage registration based on the frame-to-mosaic technique. In an other work, Xue et al. \cite{Xue2013} choose a feature based method to construct a mosaic with key frames of which positions are manually chosen. The background model is a Panoramic GMM (PGMM), extended from the model proposed by Friedman and Russell \cite{Friedman1997}. The method proposed in 2007 by Brown and Lowe \cite{Brown2007} to build a panorama is used by Xue et al. \cite{Xue2010}, Zhang et al. \cite{Zhang2010} and more recently by Avola et al. \cite{Avola2017}. This method performs a mosaic with unordered images by using a Frame-to-Mosaic approach. In the work of Sugaya and Kanatani \cite{Sugaya2005} feature points that belong to the background are selected by fitting a 2D affine space to the feature point trajectories. These features points are then used to estimate homographies by the re-normalization method \cite{Kanatani2000}. While most approaches used feature points to estimate their transformation, the method of Amri et al. \cite{Amri2010} operates on both regions and points of interest. In an other approach, Vivet et al. \cite{Vivet2009} compute the global motion with the Multiple Kernel Tracking \cite{Hager2004} method on small uniformly selected regions. This approach is computationally light and doesn't need a lot of memory. Some authors use a priori knowledge or measured data to register a pair of image as in the work of Kang et al. \cite{Kang2003} where the focal length and the size of the CCD sensor are known. When the telemetry information is available from their airborne sensor, Ali and Shah \cite{Ali2006} combined the angles with a feature based approach and a direct method. Rather than improve the image alignment, Hayman et al. \cite{Hayman2003} choose to improve the GMM proposed by Stauffer and Grimson \cite{Stauffer1999} to handle image noise and calibration errors.

\bigskip

After image registration, the last step to construct a mosaic is the blending step. It consists of mixing pixels that belong to the overlap region of images when they are warped together.

Several approaches exist, from simple ones like the triangular weighting function used by Bhat et al. \cite{Bhat2000} to more complex ones as the multi-band blending used by Xue et al. \cite{Xue2013}. In an other work, Amri et al. \cite{Amri2010} choose to use the \emph{temporal median operator}. The advantage of the temporal median scheme is that it can remove foreground from the mosaic since it supposes that moving objects doesn't stay at the same location more than half time during the initialization step. In 2005, Bevilacqua et al. \cite{Bevilacqua2005} use the alpha-update rule also known as the Infinite Impulse Response (IIR) filter used for background maintenance. In a further work, Bevilacqua and Azzari \cite{Bevilacqua2006} reduce the seam effects on the panorama by performing a tonal alignment on gray scale images. To do that, the authors use an intensity mapping function on histograms.

\bigskip

To compute the foreground detection on the current frame, it is necessary to register the image to the background.

In 2000, Bhat et al. \cite{Bhat2000} make use of the panorama building step to store information that are then used to register a new frame to the mosaic. For each frame that constitute the panorama, the pan and the tilt angles and the affine parameters are stored. The rotation angles of the new frames are used with the stored information to obtain a first coarse registration which is refined by the estimation of transformation parameters between the new frame and the rough mosaic region. In an other work, Xue et al. \cite{Xue2010} use feature points and camera parameters saved during the panoramic building step to register the current image to the panorama. A gray-level histogram is computed for the background and the current image where the pixel value distributions are previously normalized to prevent lighting changes. The Kullback-Leiber Divergence is then used to obtain the foreground probabilities of each pixel and finally thresholded to compute the foreground mask.

The image registration with a PTZ camera is a complex task because the image can be taken at the different scale from the background. To overcome this problem, Zhang et al. \cite{Zhang2010} capture images at different focal length and these images are group according to the focal length. When the current image is register to the mosaic with the feature points, the sets of feature points attached to each group of mosaic images are enlarged with the new matched feature points. In an other approach, Xue et al. \cite{Xue2013} propose a new multi-layered propagation method that cope with the number of matching features points between the current frame and the panorama that decreases when the scale of the current frame increases. A hierarchy of image at different scales is constructed where a layer groups frames taken at the same scale and layers are linked together by matching feature points. The hierarchy of layers is then used to register the current frame to the panorama by propagating correspondences through the layers. The foreground detection is computed by thresholding the minimum Mahalanobis distance between a pixel and a block centered on the corresponding background pixel. The multi-layered system is also used by Liu et al. \cite{Liu2015} but to represent the background and not to register the current frame to a panorama. Each layer is composed of a set of key frames where key frames are encoded with a spatio-temporal model. The current frame is registered with the pan, the tilt angle and the focal to find the nearest key frames and a homography is computed for the registration.

In 2008, Asif et al. \cite{Asif2008} choose to analyze the global motion by block in the image. The phase correlation is used to determinate the motion of each block which permit to obtain a first foreground estimation. Foreground blocks are divided into smaller blocks to refine the label by analyzing the sum of absolute difference for each block and their neighbors. In an other work, Ali and Shah \cite{Ali2006} suggest to use two methods to obtain foreground objects: accumulative frame differencing and background subtraction. A histogram of log-evidence is combined with the result of a hierarchical background subtraction to detect moving objects. In an recent work, Avola et al. \cite{Avola2017} propose to attach a spatio-temporal structure to each keypoints. The spatio-temporal information is used to track background feature points and label them as background or foreground. A clustering stage is also applied on keypoints to validate the foreground labeling. When two objects are represented by only one blob, because of noise or shadows, Kang et al. \cite{Kang2003} analyze the vertical projection histogram and use it to correct the segmentation.

\rowcolors{1}{}{veryLightGray}
\begin{sidewaystable*}[htbp]
	\centering
	\begin{tabular}{lllccccccc}
		\toprule
		References && Main contribution & \rot{FF} & \rot{FM} & \rot{AB} & \rot{FB} & \rot{DM} & \rot{AM} & \rot{PM}\\
		\hline
		Mittal et al.		& (2000) \cite{Mittal2000}		& Mosaic building with moving objects		& \n & \y & \n & \y & \y & \y & \n \\
		Bhat et al.			& (2000) \cite{Bhat2000}		& Mosaic building \& Background modeling	& \n & \n & \y & \n & \n & \n & \n \\
		Bartoli et al.		& (2002) \cite{Bartoli2002}		& Mosaic building							& \y & \n & \n & \y & \y & \n & \n \\
		Hayman et al.		& (2003) \cite{Hayman2003}		& Background modeling						& \y & \n & \n & \y & \n & \n & \y \\
		Kang et al.			& (2003) \cite{Kang2003}		& Reduce segmentation noise					& \n & \n & \y & \y & \n & \n & \n \\
		Bevilacqua et al.	& (2005) \cite{Bevilacqua2005}	& Mosaic building							& \y & \y & \n & \y & \n & \n & \y \\
		Sugaya et al.		& (2005) \cite{Sugaya2005}		& Mosaic building							& \n & \y & \n & \y & \n & \y & \n \\
		Ali and Shah		& (2006) \cite{Ali2006}			& Methods combined							& \y & \n & \y & \y & \y & \n & \y \\
		Bevilacqua et al.	& (2006) \cite{Bevilacqua2006}	& Tonal alignments							& \y & \y & \n & \y & \n & \n & \y \\
		Asif et al.			& (2008) \cite{Asif2008}		& Moving objects detection					& \y & \n & \n & \y & \y & \n & \n \\
		Vivet et al.		& (2009) \cite{Vivet2009}		& Mosaic building							& \n & \y & \n & \n & \n & \n & \y \\
		Amri et al.			& (2010) \cite{Amri2010}		& Temporal median operator					& \y & \n & \n & \y & \n & \n & \y \\
		Xue et al.			& (2010) \cite{Xue2010}			& Mosaic building							& \n & \y & \n & \y & \n & \n & \y \\
		Zhang et al.		& (2010) \cite{Zhang2010}		& Large zoom								& \n & \y & \n & \y & \n & \n & \y \\
		Xue et al.			& (2013) \cite{Xue2013}			& Large zoom								& \n & \y & \n & \y & \n & \n & \n \\
		Avola et al.		& (2017) \cite{Avola2017}		& Spatio-temporal keypoints	tracking		& \n & \y & \n & \y & \n & \y & \n \\
		\bottomrule
	\end{tabular}
	\caption{Panoramic methods summary. \textbf{FF}: Frame-to-Frame, \textbf{FM}: Frame-to-Mosaic, \textbf{AB}: Angle Based, \textbf{FB}: Feature Based, \textbf{DM}: Direct Method, \textbf{AM}: Affine Model, \textbf{PM}: Projective Model.}
\end{sidewaystable*}
%
\subsubsection{Dual cameras}

Instead of construct a panorama, some methods use a dual-camera system where one of the two cameras has a wide focal of view to observe the whole scene.

The camera calibration is an important step to make use of information provided by several cameras. Autocalibration is generally used contrary to calibration which necessitate some device whose the best-known example is the chessboard.

\bigskip

In 1998, Cui et al. \cite{Cui1998} need to know the relative positions and the projection model of their camera to calibrate them. Rather than using a geometry calibration which requires the relative positions between the two cameras, Chen et al. \cite{Chen2008} propose a homography calibration with polynomials without prior knowledge but at the cost of a slightly degraded mapping accuracy. In an other work, Horaud et al. \cite{Horaud2006} estimate the intrinsic parameters of both cameras and use 3D patterns for the stereo calibration. Another calibration step, named the kinematic calibration and based on the epipolar geometry, is used to rotate the PTZ camera. To achieve real-time computation, Kumar et al. \cite{Kumar2009} construct an offline look-up table with different pan and tilt angles. A neural network is then trained offline with the look-up table and the result is used to interpolate any PTZ orientation during the online image registration process. In an other work, Lim et al. \cite{Lim2003} first compute zero-positions between the static camera and the PTZ ones. The pan and tilt angles needed to track an object are derived from the projective geometry equations and image point trajectories. Several static and PTZ cameras are used in the work of Krahnstoever et al. \cite{Krahnstoever2008}. To calibrate their cameras in the same coordinate system, the authors use the \emph{foot-to-head homology} combined with a Bayesian formulation to handle measurement uncertainties \cite{Krahnstoever2005, Krahnstoever2006}.

\bigskip

Motion detection is usually performed in two step, firstly in the static camera to indicate where the moving camera has to look before it performs moving objects detection too.

\begin{figure}[!ht]
	\includegraphics[width=\columnwidth]{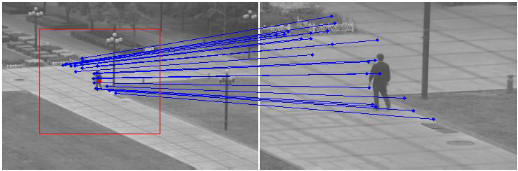}
	\caption{An example of image registration between a large-view static camera and local-view PTZ camera.\\ \textit{Source}: Images from Cui et al. \cite{Cui2014}.}
	\label{fig:dual_cameras_image_registration}
\end{figure}

In 1998, Cui et al. \cite{Cui1998} use a fish-eye camera and PTZ cameras and both kind of camera are used to monitor and track moving objects. With the fish-eye camera, the authors compute radial profiles instead of using a pixel-based background subtraction because it is more robust to shadows and small lighting changes. The tracking task is performed by a Kalman filtering. In the case of a PTZ camera, the detection and the tracking is based on the skin color. In an other work, Lim et al. \cite{Lim2003} use the method proposed by Elgammal et al. \cite{Elgammal2000} designed for stationary cameras. This method is based on non-parametric background representation which handle dynamic background and shadows. In 2014, Cui et al. \cite{Cui2014} use two cameras: a large-view static camera at low resolution and a local-view PTZ camera at high resolution. The images from the static camera are used for the background model and moving objects are detected in the images of the PTZ camera. Images are registered in three steps: a rough region is obtained with mean-shift, a 2D transformation is computed from features points with the RANdom SAmple Consensus (RANSAC) algorithm \cite{Fischler1987}, the transformation is refined with the Sum Squared Difference (SSD) method. To refine the foreground area, Horaud et al. \cite{Horaud2006} compare three aligned images.

\rowcolors{1}{}{veryLightGray}
\begin{sidewaystable*}[htbp]
	\centering
	\begin{tabular}{llccccccc}
		\toprule
		References && Main contribution & \rot{APK} & \rot{A} & \rot{FB} & \rot{DM} & \rot{AM} & \rot{PM} \\
		\hline
		Horaud et al. 		& (2006) \cite{Horaud2006}			& Calibration with epipolar geometry	& \y & \y & \n & \y & \n & \y \\
		Chen et al. 		& (2008) \cite{Chen2008}			& Two spatial mapping methods			& \n & \y & \n & \n & \n & \y \\
		Krahnstoever et al. & (2008) \cite{Krahnstoever2008}	& Combine several cameras				& \n & \y & \n & \n & \n & \n \\
		Kumar et al. 		& (2009) \cite{Kumar2009}			& Real time rectification method		& \n & \y & \y & \n & \n & \y \\
		Cui et al. 			& (2014) \cite{Cui2014}				& A three-step image registration		& \n & \n & \y & \n & \y & \n \\
		\bottomrule
	\end{tabular}
	\caption{Dual camera methods summary. \textbf{APK}: A Priori Knowledge, \textbf{A}: Autocalibration, \textbf{FB}: Feature Based, \textbf{DM}: Direct Method, \textbf{AM}: Affine Model, \textbf{PM}: Projective Model.}
\end{sidewaystable*}

%

\subsubsection{Motion compensation}
\label{sec:motion_compensation}
\begin{figure}[!ht]
	\includegraphics[width=.5\columnwidth]{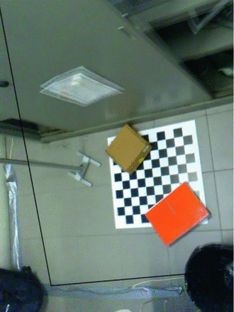}
	\includegraphics[width=.5\columnwidth]{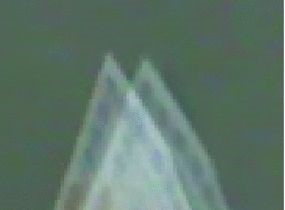}
	\caption{An example after image registration with a homography. The 2D transformation is based on the floor and we observe that the closet is misaligned. The second picture clearly shows this misalignment on the sheet paper.\\ \textit{Source}: Images from Romanoni et al. \cite{Romanoni2014}.}
	\label{fig:motion_compensation_homography_misalignment}
\end{figure}

One simplest technique to adapt the background subtraction method to a moving camera is to compensate the motion of the camera in order to realize the subtraction as in a stationary camera case. Those methods used \emph{Motion Compensation} techniques to register the current image with the background model with a 2D parametric transformation \cite{Odobez1997, Hartley2003}. After the registration step, images are configured as with a static camera and background subtraction techniques can be applied on the registered frame. Nevertheless the global estimation of the 2D transformation of the current frame with a previous one or a background model lead to foreground false alarms due to the registration errors as shown by the figure~\ref{fig:motion_compensation_homography_misalignment} and generally a refinement step is necessary.

\bigskip

Contrary to previous methods presented in section~\ref{sec:panoramic_background_subtraction}, the background model is not an extended image as a panorama but an image with the same resolution as a frame taken by the moving camera. From one frame to another the visible part of the background changes over time since the camera is moving. The background image at a time $t$ is composed of previous scene parts still visible in the camera field of view and new scene parts that appear in the current image. The background subtraction with motion compensation can also be used with a PTZ camera \cite{Murray1994, Robinault2009, Kadim2013}. In that case, instead of creating a panorama with several images, the background model has the size of a frame. This reduces the computation time and the memory allocation needed for the whole subtraction process.

To reduce errors in the final mask, some authors choose to use two models \cite{Wu2011, Wan2014, Lopez-Rubio2015, Kurnianggoro2016a, Zhao2018, Yu2019}. In 2011, Wu et al. \cite{Wu2011} compute background and foreground maps in a joint spatial-color domain with the Kernel Density Estimation (KDE) method applies on the previous pixel classification. The spatial-color cue is used with contrast and motion cues to obtain a segmentation by a Conditional Random Field (CRF) energy minimization. In an other work, Wan et al. \cite{Wan2014} construct two GMM for each feature points, based on the mean and the variance of background and foreground clusters. A foreground feature point is removed from the foreground set if its probability to belong to the foreground model is less than belong to the background model. In a recent work, Zhao et al. \cite{Zhao2018} use two confidence images: the foreground confidence image preserved the proximity captured by a GMM while the background confidence image preserves set of background spatio-temporal features. Two models are used in the work of Lopez-Rubio et al. \cite{Lopez-Rubio2015} for two different tasks: one to estimate the motion of the camera and the other one to compute the foreground. For both models, one Gaussian component represents the background and one uniform component represents the foreground. The first model is in the RGB space while the second one uses 24 features. In 2016, Kurnianggoro et al. \cite{Kurnianggoro2016a} and, in a recent work by the same authors, Yu et al. \cite{Yu2019} use a background model and a candidate background model. The candidate background guarantees that a pixel is stable on a given period before add it in the background model.

In 2014, Ferone and Maddalena \cite{Ferone2014} propose to use a neural map as background model. This map is an enlarged version of a frame where each pixel is represented by $n \times n$ weight vectors. When a pixel find a match with the background model, the corresponding neuron in the map is updated and also its neighborhood in order to take into account spatial relationship.

Image registration is done by estimating a 2D transformation between the current image and a previous one or the background model. In 1994, Murray and Basu \cite{Murray1994} use the focal length and the pan and tilt rotations given by potentiometers to estimate the position of a pixel in the previous frame. Rather than using a priori knowledge, the computation of alignment parameters can be performed with feature-based \cite{Torr1999} or direct \cite{Irani1999} methods. Generally the feature based method is preferred (see~\ref{tab:motion_compensation_methods_summary}) because it is fast to compute and the features usually used are the well-known feature points \cite{Lucas1981, Shi1994}. To save computation time and reach real-time performance, Micheloni and Foresti \cite{Micheloni2006} use a Fast Feature Selection (FFS) which select good feature points based on the quality criterion of Tomasi and a map of good feature points is maintained rather than extract features from scratch. In the case of PTZ camera, it is possible to know the intrinsic and extrinsic parameters. In an other approach, Robinault et al. \cite{Robinault2009} estimate a homography with a minimization algorithm and accelerate the computation time by using a cost function based on the location of feature points. To reject bad homography estimation, Lopez-Rubio et al. \cite{Lopez-Rubio2015} propose to find "minor errors" which occur when the model is too large or to small. A new homography is then computed based on new features points. If 10 consecutive minor errors occur then it is a severe error and the current frame is skipped. With three consecutive severe errors, both models are reset with the current frame. Since the camera is moving, some images can be blurred by the motion and this affects the accuracy of feature points detection and matching. To prevent that, Kadim et al. \cite{Kadim2013} find vertical edges compute the average absolute edge magnitude to evaluate the blurriness level of the current image and only keep images taken when the camera is in a stable position. In order to save more computational time, some authors choose to select points on a grid and track them with optical flow or with well-known track methods as the KLT \cite{Kurnianggoro2016, Kurnianggoro2016a, Minematsu2015, Minematsu2017, Yu2019}.

Feature points that belong to foreground should not be used to compute the 2D transformation and Wan et al. \cite{Wan2014} propose a two-layer iteration to estimate the transformation parameters. In the inner layer, the RANSAC algorithm is used to obtain a transformation model while in the outer layer the transformation parameters are used to classify feature points as background or foreground. The new background feature points are used to estimate a new transformation model until the classification converge. In an other work, Guillot et al. \cite{Guillot2010a} reduce matching candidates for a feature point by using a small search window to match more points.

\bigskip

In theory, after the registration step, the background model and the current frame are aligned and a foreground detection used with static camera can be applied. In practice the current frame is not perfectly aligned because of parallax generated by 3D objects that do not belong to the 2D plane described by th 2D transformation. 

A common way to handle the parallax is to use the neighborhood of a pixel to classify. In 1997, Odobez and Boutemy \cite{Odobez1997} use only motion measurements rather than intensity change measurements. These measurements are embedded in a multiscale Markov Random Field (MRF) framework to encourage neighboring pixels to have the same label. A voting technique is proposed by Paragios and Tziritas \cite{Paragios1999} to choose the regularization parameter of the cost function to minimize to obtain a binary mask. In an other work, Ren et al. \cite{Ren2003} propose a Spatial Distribution of Gaussians (SDG) model to provide a temporal and spatial distribution of the background where the authors assume that the intensity distribution of each pixel can be modeled by a two-component MOG. The methods proposed by Kim et al. \cite{Kim2013} and Viswanath et al. \cite{Viswanath2015} compared the intensity of a pixel labeled as foreground and the intensities of its neighborhood in the background model. A low difference between intensities means a false alarm but the silhouette of a moving object can be affected by this refinement. Kim et al. used PID control-based tracking and probabilistic morphology refinement step to recover the silhouette. In the approach proposed by Romanoni et al. \cite{Romanoni2014}, two histograms are computed: one on the neighborhood of a pixel and another one based on the neighborhood and the intensities history of the same pixel. The Bhattacharyya distance is used with a threshold to detect moving objects. In an other work, Minematsu et al. \cite{Minematsu2015} proposed to find an intensity match between a pixel and another one in a search region. This region represents the neighborhood of a pixel where the size of the region depends on re-projection errors. Later, the authors proposed to update the background model by selecting background pixels based on a similarity measure and the re-projection error. Instead of building and maintaining a background model, Kadim et al. \cite{Kadim2013} choose to detect moving objects by using successive frames. The Wronskian detector \cite{Durucan2001} is used to detect moving objects between the current and the previous frame. The authors also use the neighborhood to refine their motion map and they remove false moving blobs by validating only those that are detected for at least two successive frames. More recently, Zhao et al. \cite{Zhao2018} work with superpixel at different level. A competition between background and foreground cues is organized. The result gives the classification of the corresponding superpixel. To counteract error alignment accumulations, a strong updating strategy is applied on background pixels. In 2019, Yu et al. \cite{Yu2019} align the two previous frames to the current one and to save computation time they compute the frame difference on the average on the pixel and its 8-neighborhood. To remove shadows from the foreground, the consistency of local changes is checked. A consistency points out a shadow area while there is no consistency for a moving object. In addition, a lighting influence threshold is used to managed illumination changes in the entire frame.

In the case where the application domain is constrained, the segmentation of the scene can be an additional information to moving object detection. Perera et al. \cite{Perera2006} and Huang et al. \cite{Huang2010} both work on aerial images and try to segment vehicles on roads. Perera et al. \cite{Perera2006} choose to use scene understanding to segment the image into region and attribute a predefined class, as road or tree, to each region. Huang et al. \cite{Huang2010} segment images into regions and road regions are identified by the size and the straight line property of the region contour. In both methods, a region is a moving object according to its position relative to a road region. Perera et al. also use the scene understanding to remove feature points on trees to obtain a better estimation of the homography. As for Huang et al. combine the result of image segmentation with the one of frame difference to obtain a better foreground segmentation.

Since the camera is moving, some parts of the scene disappear while others appear. Parts that disappear do not need special treatment and they are just remove when the background is updated. However, new parts have to be integrated in the classification and in some methods, they are initialized as background \cite{Kim2013}. In their method, Lopez-Rubio et al. \cite{Lopez-Rubio2015}, find the closest labeled pixel of a new one. If the new pixel belongs to the background model of its closest neighbor, then this background model is used as initialization, otherwise with a neutral state.

A traditional approach to reduce noise in the binary mask is morphological operation (see~\ref{tab:motion_compensation_methods_summary}). This technique can remove small groups of pixels falsely labeled as foreground and fill small holes in the foreground segmentation. To remove noise pixels connected to foreground, Solehah et al. \cite{Solehah2012} propose to compare the histogram of the current image with the one of the warped background and threshold it to re-classify the pixels.

\rowcolors{1}{}{veryLightGray}
\begin{sidewaystable*}[htbp]
	\centering
	\begin{tabular}{lllccccc}
		\toprule
		References && Main contribution & \rot{FB} & \rot{DM} & \rot{AM} & \rot{PM} & \rot{MF} \\
		\hline
		Murray and Basu			& (1994) \cite{Murray1994}			& Real time motion detection							& \n & \n & \n & \y & \y \\
		Odobez and Bouthemy		& (1997) \cite{Odobez1997}			& Statistical regularization framework					& \n & \y & \y & \n & \n \\
		Paragios and Tziritas	& (1999) \cite{Paragios1999}		& Regularization parameter by a voting technique		& \n & \y & \y & \n & \n \\
		Ren et al.				& (2003) \cite{Ren2003}				& Spatial distribution of Gaussians						& \y & \n & \y & \n & \y \\
		Micheloni and Foresti	& (2006) \cite{Micheloni2006}		& Real time												& \y & \n & \n & \n & \n \\
		Perera et al.			& (2006) \cite{Perera2006}			& Use scene understanding								& \y & \n & \n & \y & \y \\
		Robinault et al.		& (2009) \cite{Robinault2009}		& Real time												& \y & \y & \n & \y & \n \\
		Guillot et al.			& (2010) \cite{Guillot2010a}		& Feature points matching								& \y & \n & \n & \y & \n \\
		Huang et al.			& (2010) \cite{Huang2010}			& Combine frame difference and image segmentation		& \y & \n & \n & \y & \n \\
		Wu et al.				& (2011) \cite{Wu2011}				& Spatial-color cue for CRF								& \y & \n & \n & \y & \n \\
		Solehah et al.			& (2012) \cite{Solehah2012}			& Refine foreground with local histogram processing		& \y & \n & \n & \y & \y \\
		Kadim et al.			& (2013) \cite{Kadim2013} 			& Avoid blurred images									& \y & \n & \n & \y & \n \\
		Kim et al.				& (2013) \cite{Kim2013}				& Spatio-temporal update scheme							& \y & \n & \n & \y & \y \\
		Ferone and Maddalena	& (2014) \cite{Ferone2014}			& Self-organizing background subtraction				& \y & \n & \n & \y & \n \\
		Romanoni et al.			& (2014) \cite{Romanoni2014}		& Temporal + Spatio-Temporal Histograms algorithm		& \y & \n & \n & \y & \y \\
		Wan et al.				& (2014) \cite{Wan2014}				& Two-layer iteration									& \y & \n & \y & \n & \n \\
		Lopez-Rubio et al.		& (2015) \cite{Lopez-Rubio2015}		& Two probabilistic models								& \y & \n & \n & \y & \n \\
		Minematsu et al.		& (2015) \cite{Minematsu2015}		& Re-projection error									& \y & \n & \n & \y & \n \\
		Viswanath et al.		& (2015) \cite{Viswanath2015}		& Spatio-temporal Gaussian model						& \y & \n & \n & \y & \n \\
		Kurnianggoro et al.		& (2016) \cite{Kurnianggoro2016}	& Using dense optical flow								& \y & \n & \n & \y & \n \\
		Kurnianggoro et al.		& (2016) \cite{Kurnianggoro2016a}	& Candidate background model							& \y & \n & \n & \y & \y \\
		Minematsu et al.		& (2017) \cite{Minematsu2017}		& Improved updating background models					& \y & \n & \n & \y & \y \\
		Zhao et al.				& (2018) \cite{Zhao2018}			& Integration of foreground and background cues			& \y & \n & \n & \y & \n \\
		Yu et al.				& (2019) \cite{Yu2019}				& Improve background subtraction						& \y & \n & \n & \y & \y \\
		\bottomrule
	\end{tabular}
	\caption{Motion compensation methods summary. \textbf{FB}: Feature Based, \textbf{DM}: Direct Method, \textbf{AM}: Affine Model, \textbf{PM}: Projective Model, \textbf{MF}: Morphological Filtering.}
	\label{tab:motion_compensation_methods_summary}
\end{sidewaystable*}

%
\subsubsection{Subspace segmentation}
\label{sec:subspace_segmentation}
In this section, moving objects detection methods use the trajectories of feature points to separate the background and the foreground. Contrary to the previous approaches, there is no registration between images to apply a background subtraction technique. The features points are labeled according to the analysis of their trajectories and the label information is propagated to the whole image to obtain a pixel-wise segmentation.

\begin{figure}[!ht]
	\includegraphics[width=\columnwidth]{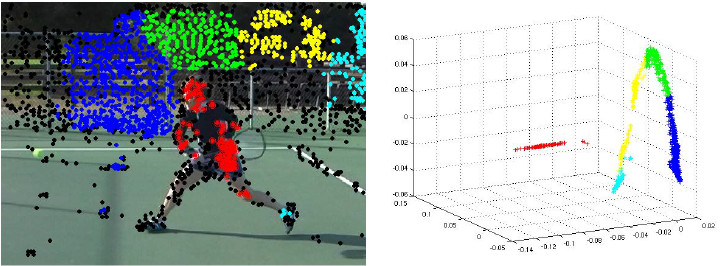}
	\caption{An example of clustering trajectories into a subspace (right) and the result on the image (left).\\ \textit{Source}: Images from Elqursh and Elgammal et al. \cite{Elqursh2012}.}
	\label{fig:subspace_segmentation_trajectories_clustering}
\end{figure}

\bigskip

In 2009, Sheikh et al. \cite{Sheikh2009} use three long term trajectories to construct a 3D subspace. Feature points whose trajectories belong to this subspace are considered as part of the background while the others are foreground. In the proposed method of Elqursh and Elgammal \cite{Elqursh2012} a subspace is constructed with trajectory affinities computed on motion and spatial location. The trajectories in the embedded subspace are then clustered and labeled foreground or background by minimizing an energy function which combine multiple cues. The result of this segmentation is presented in the figure~\ref{fig:subspace_segmentation_trajectories_clustering}. In an other work, Nonaka et al. \cite{Nonaka2013} cluster the trajectories by using three different distances and label the cluster based on the shape and the size. To reduce the computation time and the memory resource, the trajectories from two consecutive frames are used rather than long term trajectories. In 2014, Berger and Seversky \cite{Berger2014} managed the changing number of trajectories over time by a dynamic subspace tracking. At each frame, the camera parameters are updated and used to update the shape of the trajectories. More recently, Sajid et al. \cite{Sajid2019} propose to combine motion and appearance. The motion module performs a low-rank approximation of the background dense motion with an iterative method. The probability of each pixel belongs to the foreground is estimated from the pixel-wise motion error between the background motion approximation and the one observed. The appearance module models background and foreground with GMM.

\bigskip

In order to obtain a binary mask, the sparse label information is propagated to the whole image. The common method to propagate the information is to segment the image by constructing a pairwise MRF and minimizing the energy generally with the graph-cut algorithm. A pairwise MRF is a graph where vertices represent the pixels and the edges connect the vertices with their neighborhood as a grid structure over the image. The energy of a MRF is composed of two terms: the unary term and the binary term. The unary term is used to assign a label to a vertex while the binary term encourages to assign the same label to vertices connected by an edge in order to smooth the segmentation. A cut is then found in the graph by minimizing the energy to obtain an image segmentation.

In 2009, Sheikh et al.\cite{Sheikh2009} use the kernel density estimation method to obtain two models, one for the background and one for the foreground. The graph-cut algorithm is then used to minimize an energy function on a MRF. In the method of Elqursh and Elgammal \cite{Elqursh2012} the motion model is propagated to each pixel with a pairwise MRF and estimate the labels with a Bayesian filtering. In an other approach, Nonaka et al. \cite{Nonaka2013} propose to use a case database, which described the foreground with the color and the location, in the segmentation step for the next frame.

\rowcolors{1}{}{veryLightGray}
\begin{sidewaystable*}[htbp]
	\centering
	\begin{tabular}{llcccccc}
		\toprule
		References && Main contribution & \rot{TTL} & \rot{FB} & \rot{DOP} & \rot{MRF} & \rot{GCA} \\
		\hline
		Sheikh et al.			& (2009) \cite{Sheikh2009}		& Three dimensional subspace	& \y & \y & \n & \y & \y \\
		Elqursh and Elgammal	& (2012) \cite{Elqursh2012}		& Appearance and motion models	& \n & \y & \n & \y & \y \\
		Nonaka et al.		 	& (2013) \cite{Nonaka2013}		& Reduce time computation		& \n & \y & \n & \y & \y \\
		Berger and Seversky		& (2014) \cite{Berger2014}		& Dynamic subspace tracking		& \y & \y & \n & \y & \y \\
		Sajid et al.			& (2019) \cite{Sajid2019}		& Combine motion and appearance	& \n & \n & \y & \y & \y \\
		\bottomrule
	\end{tabular}
	\caption{Subspace segmentation methods summary. \textbf{LTT}: Long Term Trajectory, \textbf{FB}: Feature Based, \textbf{DOP}: Dense Optical Flow, \textbf{MRF} : Markov Random Field, \textbf{GCA}: Graph Cut Algorithm}
\end{sidewaystable*}

%
\subsubsection{Motion segmentation}
\label{sec:motion_segmentation}
\begin{figure}[!ht]
	\centering
	\includegraphics[width=0.8\columnwidth]{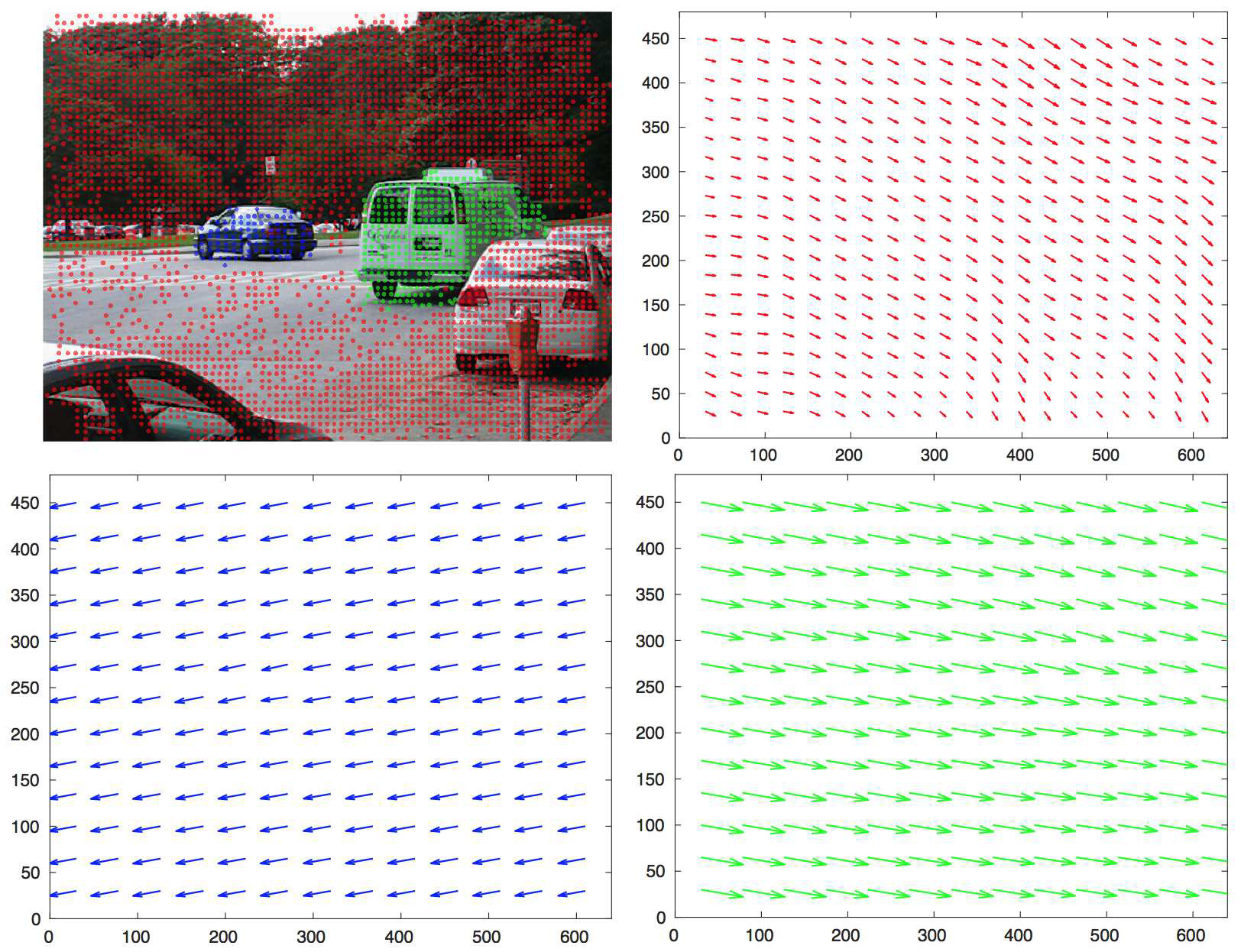}
	\caption{An example motion segmentation on the top left image. The three other images represent the optical flow of the three motions observed in the image.\\ \textit{Source}: Images from Zhu and Elgammal \cite{Zhu2017}.}
	\label{fig:motion_segmentation}
\end{figure}

The same way as the previous section, the methods presented here uses the trajectories of the feature points to segment each frame of the video as static or moving but without using a subspace (see figure~\ref{fig:motion_segmentation}). Those methods are inspired by the methods called Motion Segmentation in the literature which segment the image according to the apparent motions. Here the methods presented go further than just segment each frame of the video by the 2D motions by proposing a background/foreground labeling.

\bigskip

In 2015, Yin et al. \cite{Yin2015} cluster feature points according to their trajectory similarity and reject false trajectories by using the PCA algorithm. In an other work, Bideau et al. \cite{Bideau2016} use the translational flow obtained by the subtraction of the dense optical flow and the rotational flow. The angle field is then estimated from the translational flow according to the magnitude which indicates the reliability of the flow angle. Then the conditional flow angle likelihood estimate the probability that the flow direction of a pixel corresponds to the one estimated. Finally, the Bayes' rule is used to obtain the posterior probability for each pixel which is used for the final segmentation. The authors also proposed to segment the first frame of the video by choosing three superpixels with a modified RANSAC algorithm in order to estimate the motion of the background. In an other approach, Kao et al. \cite{Kao2016} recover the 3D motions from the 2D motions observed by using motion vanishing point and the estimated depth of the scene. The final segmentation is applied on the 3D motions. The method proposed by Zhu and Elgammal \cite{Zhu2017} first clusters trajectories based on their affinities and propagate the label of trajectories dynamically. The clusters automatically adapt to the number of foreground object in the frames by computing intra-cluster variation. In a recent work, Sugimura et al. \cite{Sugimura2018} use the OneCut algorithm to segment frames. Rather than manually select seeds by hand for the OneCut segmentation, the authors propose to find automatically the seeds by using motion boundaries computed by the Canny detector on the magnitude and direction flow fields. Foreground seeds are selected inside enclosed motion boundaries while background seeds are selected on rectangles that enclose motions boundaries. Recently, Huang et al. \cite{Huang2019} estimate a dense optical flow by using FlowNet2.0 \cite{Ilg2016} an optical flow estimation algorithm with deep networks. The background optical flow is estimated by a quadratic transformation function with the Constrained RANSAC Algorithm (CRA). The CRA is a modified version of the RANSAC algorithm to avoid overfitting and improving the searching efficiency.

\bigskip

As in the previous section~\ref{sec:subspace_segmentation}, sparse labeling information is propagated to the whole image to obtain a dense labeling.

In 2015, Yin et al. \cite{Yin2015} propose a trajectory-controlled watershed segmentation algorithm to propagate the label information. After applying a bilateral filtering to smooth the image and enhance the edges, gradient minima and the trajectory points are selected as markers. Those markers are used by the watershed algorithm as seeds to obtain a segmentation for which the regions are labeled background or foreground according to the labels of the trajectories. Finally, the background/foreground information is propagated to the unlabeled regions by minimizing an energy function on a MFR with the graph-cut algorithm. The Multi-Layer Background Subtraction (MLBS) proposed by Zhu et al. \cite{Zhu2017} propose a multi-label segmentation rather than a binary segmentation. Each motion cluster is associated to a layer. For each layer, a pixel-wise motion estimation is performed by a Gaussian Belief Propagation (GaBP). Then the appearance model and the prior probability map are updated with the motion estimation and they are used to compute the posterior probability map. The multi-label segmentation is performed on the posterior probability map by the minimization of the energy of a pairwise MRF. In an recent work, Sugimura et al. \cite{Sugimura2018} prevent unreliable magnitude and direction foreground flow field by introducing a prediction based on the lasts foreground estimated regions. In the case where the magnitude and direction foreground are unreliable, the prediction is used rather than the two flow fields as the segmentation result otherwise the prediction is jointly used with the two others flow fields. The OneCut is applied a second time with the appearance information in order to improve the final segmentation. In an other work, Kao et al. \cite{Kao2016} obtain a binary mask by segmenting the 3D motions with three different clustering methods: simple k-means clustering, spectral clustering with a 4-connected graph and with a fully connected graph. Recently, Huang et al. \cite{Huang2019} propose a dual judgment mechanism to separate the foreground from the background. The foreground is estimated by thresholding the difference of the estimated background optical flow and the one estimated by FlowNet2.0. In order to take into account the case where the camera is zooming, a second judge mechanism is based on thresholding the difference of cosine angles.

\rowcolors{1}{}{veryLightGray}
\begin{sidewaystable*}[htbp]
	\centering
	\begin{tabular}{llcccc}
		\toprule
		References && Main contribution & \rot{TTL} & \rot{FB} & \rot{DOP} \\
		\hline
		Yin et al.			& (2015) \cite{Yin2015}			& Trajectoy-controlled watershed segmentation	& \y & \y & \n \\
		Bideau et al.		& (2016) \cite{Bideau2016}		& Combine angle and magnitude					& \n & \n & \y \\
		Kao et al.			& (2016) \cite{Kao2016}			& 3D motions segmentation						& \n & \n & \y \\
		Zhu and Elgammal	& (2017) \cite{Zhu2017}			& Multi-label background subtraction			& \n & \y & \n \\
		Sugimura et al.		& (2018) \cite{Sugimura2018}	& Automatic OneCut method						& \n & \y & \y \\
		Huang et al.		& (2019) \cite{Huang2019}		& Dual judgment mechanism						& \n & \n & \y \\
		\bottomrule
	\end{tabular}
	\caption{Motion segmentation methods summary. \textbf{LTT}: Long Term Trajectory, \textbf{FB}: Feature Based, \textbf{DOP}: Dense Optical Flow}
\end{sidewaystable*}
%

\subsection{Several parts}
Approximate the scene with one plane limits the environment to be simple or far away from the camera. In order to handle complex scenes, with high depth variations, techniques were developed to approximate the scene by several planes.

%

\subsubsection{Plane+Parallax}
The Plane+Parallax decomposition is a scene-centered representation \cite{Irani2002}. As in the previous section, this technique firstly compensates the camera motion with a 2D parametric transformation that describes the dominant plane in the scene. After the registration process, camera rotation and zoom are eliminated and misaligned pixels correspond either to the parallax caused by the camera translation or to a moving object. Then, residual displacements belong to the scene form a radial field centered at the epipole \cite{Irani1997}.

\begin{figure}[!ht]
	\centering
	\includegraphics[width=0.5\columnwidth]{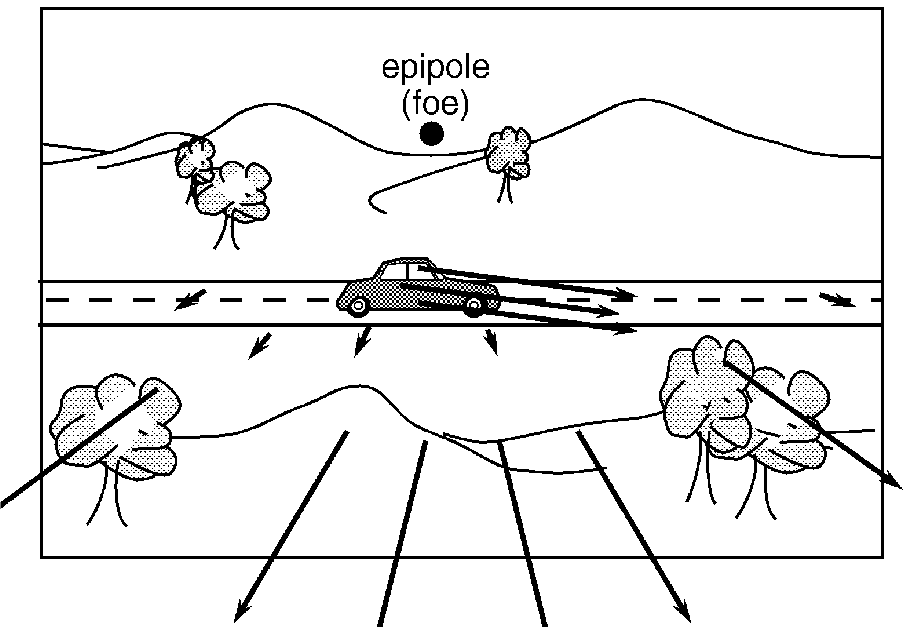}
	\includegraphics[width=0.5\columnwidth]{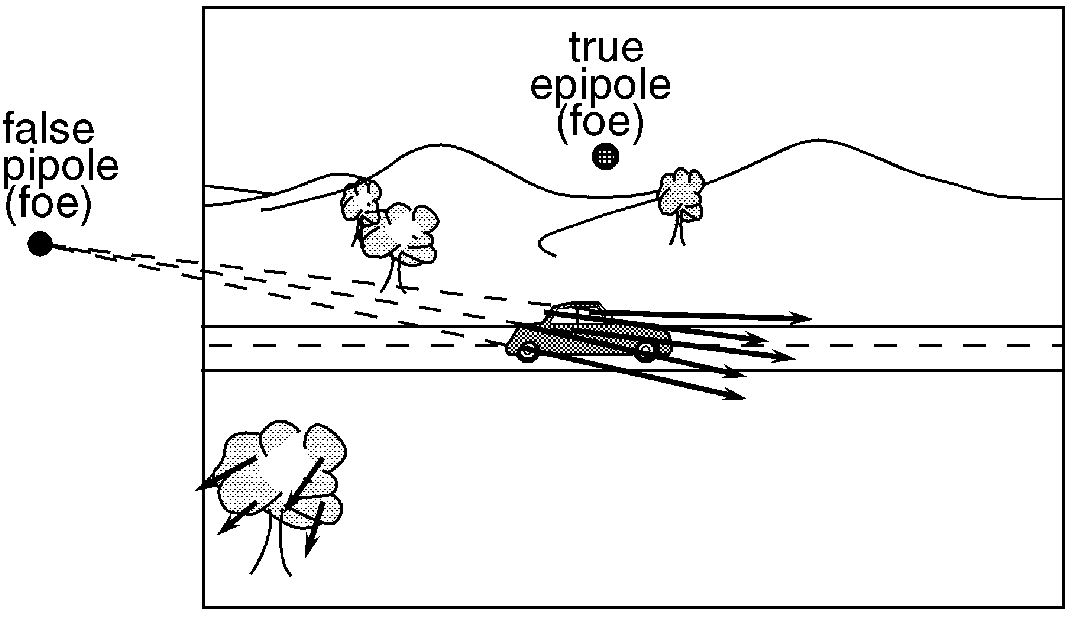}
	\caption{An illustration.\\ \textit{Source}: Images from Irani and Anandan \cite{Irani1998}.}
	\label{fig:plane_plus_parallax}
\end{figure}

In 1998, Irani and Anandan \cite{Irani1998} stratify the moving object detection problem and propose a method that handles from 2D scenes up to 3D complex scenes. The first level of the stratification is the approximation of the scene by a 2D plane. A single 2D parametric transformation is estimated between two images and used to warp them. Misalignments correspond to moving objects. The second level handle misalignments due to the parallax. Several 2D planes are estimated successively with the same method in the previous level and regions which are inconsistent with the motion of any 2D planes are moving objects. When the scene is complex, with many small moving objects are different depths, the two previous methods cannot correctly make the detection. In this case, the third level with a Plane+Parallax scene representation is used. The authors noticed that the residual movements after the registration are due to the translation motion of the camera and they form a radial field centered at the Field Of Expansion (FOE). The estimation of the FOE can be used to apply the \emph{Epipolar Constraint} but the estimation can be biased by moving objects as shown in the figure~\ref{fig:plane_plus_parallax}. To avoid this, the authors proposed a \emph{Parallax-Based Rigidity Constraint} which is a consistency measure between two points over three consecutive frames. One of the two point is known static in order to evaluate the label of the second point. In an other work, Sawhney et al. \cite{Sawhney1999} impose the \emph{Shape Constancy} and the epipolar constraint over several frames to estimate a robust image alignment. The authors used the Plane+Parallax decomposition to enforce the two constraints.

In 2005, Kang et al. \cite{Kang2005} use the consistency constraint. The advantages of this constraint are: the reference plane does not need to be the same. It could be the floor and then a wall for example. Static points are not necessary and the assumption of small camera displacement between two consecutive frames are not required. The authors combined the epipolar constraint and a structure consistency constraint to eliminate false detections due to the parallax. From the epipolar constraint an angular difference map is created and from the structure consistency constraint a depth variation map is created for each residual pixel. Rather than propose a binary mask, a likelihood map is computed on a sliding window and used directly by a tracking algorithm.

There exist one particular case where the Plane+Parallax methods do not work: when the camera and an object both move in the same direction with constant velocities. The constraints defined to distinguish the parallax and a moving object are verified and the object is labeled static.

\rowcolors{1}{}{veryLightGray}
\begin{sidewaystable*}[htbp]
	\centering
	\begin{tabular}{llccccc}
		\toprule
		References && Main contribution & \rot{FB} & \rot{DM} & \rot{AM} & \rot{PM} \\
		\hline
		Irani and Anandan	& (1998) \cite{Irani1998}	& Handle 2D and 3D scenes					& \n & \y & \n & \y \\
		Sawhney et al. 		& (1999) \cite{Sawhney1999}	& Shape constancy and epipolar constraint	& \y & \y & \n & \y \\
		Kang et al. 		& (2005) \cite{Kang2005}	& Structure consistency and angular map		& \y & \n & \n & \y \\
		\bottomrule
	\end{tabular}
	\caption{Plane+Parallax methods summary. \textbf{FB}: Feature Based, \textbf{DM}: Direct Method, \textbf{AM}: Affine Model, \textbf{PM}: Projective Model}
\end{sidewaystable*}
%

\subsubsection{Multi planes}
Multi planes scene representation was firstly used in motion segmentation \cite{Darrell1991,Wang1994,Ayer1995}.

Contrary to Motion Compensation method where only one image alignment is computed, several alignments are estimated in the case of multi-layers approaches. Cascade of RANSAC is a very used technique to estimate several real planes in a scene \cite{Jin2008,Patwardhan2008,Zhang2012a,Zamalieva2014a,Hu2015,Zhou2017}. Here is the general principle: RANSAC is used on feature points to estimate one 2D transformation between two images in the video sequence. Feature points that fit the homography are removed from the process and a new transformation is estimated with the residual feature points. This process is repeated until a condition is reached.

\begin{figure}[!ht]
	\centering
	\includegraphics[width=0.4\columnwidth]{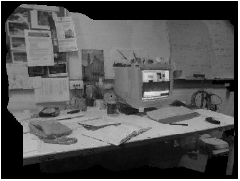}
	\includegraphics[width=0.4\columnwidth]{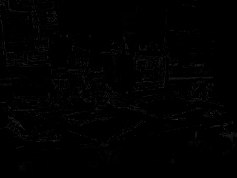}\\
	\includegraphics[width=0.4\columnwidth]{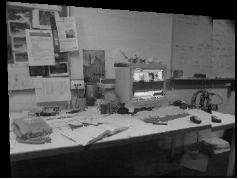}
	\includegraphics[width=0.4\columnwidth]{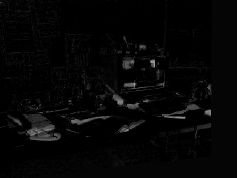}
	\caption{An example of image registration with several planes (top line) compared with image registration with one plane (bottom line). The left column represents the rectified frame after the compensation and the right column represents the disparity.\\ \textit{Source}: Images from Jin et al. \cite{Jin2008}.}
	\label{fig:several_planes_registration}
\end{figure}

In 2008, Patwardhan et al. \cite{Patwardhan2008} use a training step to automatically initialize the number of layers. Layers are estimated iteratively on color of pixels by Sampling-Expectation refining process. The method of Zhang et al. \cite{Zhang2012a} propose to adaptively adjust the parameters of RANSAC to handle simple and complex classes of scenes. Feature points are hierarchically clustered based on Euclidean distance criterion on optical flow data. A cluster is labeled as background if it has a widespread spatial distribution. Then, the number of layers is estimated iteratively by increasing the number layers until a consensus is reached. In an other work, Zamalieva et al. \cite{Zamalieva2014a} modify the GRIC score to find out if the scene can be approximated by one plane or by several planes. The modified GRIC score is computed on one homography or on the fundamental matrix. If the homography wins, it is chosen to compensate the camera motion. On the other case, a cascade of RANSAC is used to compute several homographies. In the approach of Hu et al. \cite{Hu2015} feature points are first classified as background or foreground and use them to compensate the camera motion by a homography. The authors use one plane for the frame compensation but they approximate the scene by several planes during the feature points classification by computing the fundamental matrix and using the epipolar constraint. In an other approach, Kim et al. \cite{Kim2016} estimate several homographies by clustering trajectories into the Distance and Motion Coordinate (DMC) system. From the biggest clusters, two regression lines are derived and used to find the preliminary background clusters. Homographies are estimated with the RANSAC algorithm from those background trajectories after another clustering step. Rather than find real planes in the scene, Zamalieva et al. \cite{Zamalieva2014b} propose to create parallel hypothetical planes based on the dominant plane in the scene. These planes are estimated with the vanishing line and the vertical vanishing point. The image registration is computed by homographies estimated for each hypothetical plane.

\bigskip

When several homographies are used to register the background, it is necessary to find which homography have to be applied for each pixel. In both work of Jin et al. \cite{Jin2008} and Zamalieva et al. \cite{Zamalieva2014a} pixel intensity similarity is computed for each homography to select a plane for the candidate pixel. In 2008, Jin et al. \cite{Jin2008} assign non-overlap pixels to layers with \emph{Minimal Span Tree} to represent scene smoothness. In 2014, Zamalieva et al. \cite{Zamalieva2014a} handle occluded background pixels by performing a majority voting on neighbor pixels associated to a plane.

\bigskip

Foreground detection step is very close to those used for static camera thanks to the image registration step \cite{Jin2008,Zhang2012a}. Jin et al. \cite{Jin2008} use mixture of Gaussians and a background panorama to detect moving objects while Zhang et al. \cite{Zhang2012a} simply assign a pixel to background based on intensity difference thresholding. In an other work, Patwardhan et al. \cite{Patwardhan2008} assign pixels to one layer in the training stack or identifies them as foreground. Spatio-temporal subvolume identify candidate layers and non-parametric KDE is used to estimate the probability that the current pixel belongs to each candidate layers. In a recent work, Zhou et al. \cite{Zhou2017} detect regions that became visible by motion parallax and produce false alarms. The authors combine these regions information with a codebook-based background segmentation.

\rowcolors{1}{}{veryLightGray}
\begin{sidewaystable*}[htbp]
	\centering
	\begin{tabular}{lllcccc}
		\toprule
		References && Main contribution & \rot{RP} & \rot{IP} & \rot{CR} & \rot{EG} \\
		\hline		
		Jin et al.			& (2008) \cite{Jin2008}				& Cascade of RANSAC						& \y & \n & \y & \n \\
		Patwardhan et al.	& (2008) \cite{Patwardhan2008}		& Training stack of layers				& \y & \n & \n & \n \\
		Zhang et al.		& (2012) \cite{Zhang2012a}			& Multi-classes RANSAC 					& \y & \n & \y & \n \\
		Zamalieva et al.	& (2014) \cite{Zamalieva2014a}		& Adaptive motion compensation			& \y & \n & \y & \y \\
		Zamalieva et al.	& (2014) \cite{Zamalieva2014b}		& Stack of hypothetical 3D planes		& \n & \y & \n & \y \\
		Hu et al. 			& (2015) \cite{Hu2015}				& Epipolar geometry						& \y & \n & \n & \y \\
		Kim et al.			& (2016) \cite{Kim2016}				& Distance and Motion Coordinate system	& \y & \n & \n & \n \\
		Zhou et al.			& (2017) \cite{Zhou2017}			& Regions revealed by motion parallax	& \y & \n & \y & \n \\
		\bottomrule
	\end{tabular}
	\caption{Multi layers methods summary. \textbf{RP}: Real Planes, \textbf{IP}: Imaginary Planes, \textbf{CR}: Cascade of RANSAC, \textbf{EG}: Epipolar Geometry.}
\end{sidewaystable*}
%

\subsubsection{Split image in blocks}

In the literature, one identifies two ways to divide an image into blocks. The first one simply divides the image into a regular grid where each block has a predefined size. The second technique uses superpixel segmentation methods. Each block represents a region in the image whose features depend on the segmentation method.

\begin{figure}[!ht]
	\centering
	\includegraphics[width=\columnwidth]{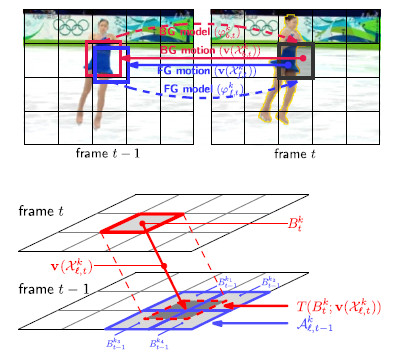}
	\caption{An example of a technique to divide the image into a regular grid and compensate the motion by blocks.\\ \textit{Source}: Images from Lim et al. \cite{Lim2012}.}
	\label{fig:split_image_in_blocks}
\end{figure}

In some methods \cite{Yi2013,Yun2015,Chung2016,Yun2017} the motion compensation is estimated on the whole image as in the section~\ref{sec:motion_compensation} but others compensate the camera motion by blocks. Rather than compute one homography for the whole image, one homography for each grid cell could be computed to register images \cite{Sun2016,Wu2017}. Some authors propose to estimate two types of motion for each block: one for the background and one for the foreground \cite{Kwak2011,Lim2012,Lim2014}. In 2011, Kwak et al. \cite{Kwak2011} choose to use non-parametric Belief Propagation to reduce the noise in optical flow and recover the missing background motion. To estimate background and foreground motion, Lim et al. \cite{Lim2012} simply use sparse optical flow. In an other method, Kim et al. \cite{Kim2013a} propose a multi-resolution motion propagation to compensate the camera motion on blocks. If a block does not have background feature points to estimate its transformation, the parameters are propagated from the blocks at a higher level. In the method of Lim and Han \cite{Lim2014}, the previous segmentation mask is warped with dense optical flow and use the warped mask to compute dense motion for background and foreground independently. In 2016, Sun et al. \cite{Sun2016} compute two kinds of motion. The first one is computed over a regular grid with the \emph{As Similar As Possible} method. The motion of the whole image is a set of homographies. The second motion is computed over superpixels with the KLT technique. These two motions are then used to obtain a background/foreground segmentation from motions.

\bigskip

The blocks are also used to model the scene. Rather than model each pixel in the image, each block is represented by one model which reduce computation time.

In 2013, Yi et al. \cite{Yi2013} choose to model each block with a Single Gaussian Model (SGM). After motion compensation, one block generally overlap several blocks in the previous frame. In order to update block models in the current frame, the overlap block models are mixed together where each block is weighted proportionally to the overlapping area. The same mixing blocks is used by Lim et al. \cite{Lim2012} for their temporal model propagation step and they additionally use a spatial step to enforce the spatial coherence. The methods of Kwak et al. \cite{Kwak2011} and Lim and Han \cite{Lim2014} also combine motion and appearance models. In 2015, Yun and Choi \cite{Yun2015} propose to improve the method of Yi et al. \cite{Yi2013} with a selectively update step based on a sampling map. Only some pixels are chosen according to temporal and spatial properties to update the model. In a further work, Chung et al. \cite{Chung2016} regulate the background model of Yi et al. \cite{Yi2013} by including foreground cues coming from frame differencing.

\bigskip

Once the models are updated, the data are combined together to create the final segmentation mask for the current frame.

In 2011, Kwak et al. \cite{Kwak2011} predict the appearance model of each block by a weighted sum of Gaussian-blurred blocks of the previous frame. In order to reduce segmentation errors, some methods \cite{Kwak2011,Lim2012,Lim2014,Chung2016} propose to iterate the process on motion and appearance models until the models converge. The method of Lim et al. \cite{Lim2012} and the one of Lim and Han \cite{Lim2014} both iterate on motion and appearance estimations to obtain a segmentation mask at each frame. In their approach, Lim and Han \cite{Lim2014} choose to use superpixel rather than a grid because this kind of pixel groups has color and motion consistency. In an other work, Yi et al.\cite{Yi2013} use two background models with ages to reduce foreground and noise contamination. Models are swapped when the candidate model is older than the current model and the new candidate model is initialized to remove contaminations. In 2017, Makino et al. \cite{Makino2017} use the method of Yi et al. \cite{Yi2013} as a baseline to compute an anomaly score map. The authors also compute a motion score map based on optical flow angles after motion compensation. The two score maps are merged in the moving object detection step. In order to manage slow moving objects, Yun et al. \cite{Yun2017} update the SGM block-based model of Yi et al. \cite{Yi2013} according to the foreground velocity. In the case where the foreground moves less than a block size during several frames, the SGM mean is updated with the illumination change and the average intensity of the block. The SGM variance is increased according to the current block intensity and the mean the previous and current time. The authors also reduce false positives by combining threshold labeling and watershed segmentation. In an other work, Kim et al. \cite{Kim2013a} combine sparse optical flow clustering with the Delaunay triangulation method in order to complete the missing detection information of the Frame Differencing method. The optical flow clustering is computed on blocks with the K-means method. In an other approach, Sun et al. \cite{Sun2016} create two segmentations, one from motion and one from appearance. The motion one is created from the difference between the camera motion estimation and the superpixels motion estimation. Identical motions on superpixels come from the background and they are used as seeds for a region growing propagation. The appearance segmentation is based on color and Local Binary Similarity Patterns (LBSP). The two segmentations are then combined with MRF and the final segmentation is obtained by graph-cut. In a recent work, Wu et al. \cite{Wu2017} use a coarse-to-fine method to detect foreground objects. Each block of the regular grid is warped according to its dominant motion over a sliding window. The Mean Squared Error (MSE) is then used as a threshold to obtain a coarse foreground region. The motion of the coarse foreground region is decomposed into background and foreground motions thanks to inpainting method. The fine foreground is obtained by an adaptive thresholding method. After compensating the camera motion by a Hierarchical Block-Matching algorithm, Szolgay et al. \cite{Szolgay2011} build a Modified Error Image (MEI) from the result of the frame difference. A spatio-temporal background Probability Density Function (PDF) for each pixel of the MEI is computed with the Kernel Density Estimation (KDE). Pixels are then labeled as background or foreground according to the PDFs and pixels are finally clustered with their motion, color and location.

\rowcolors{1}{}{veryLightGray}
\begin{sidewaystable*}[htbp]
	\centering
	\begin{tabular}{lllccccccc}
		\toprule
		References && Main contribution & \rot{RG} & \rot{S} & \rot{MCI} & \rot{MCB} & \rot{BFM} & \rot{MAM} & \rot{IM} \\
		\hline
		Kwak et al.			& (2011) \cite{Kwak2011}		& Hybrid inference motion/appearance					& \y & \n & \n & \y & \y & \y & \y \\
		Szolgay et al.		& (2011) \cite{Szolgay2011}		& Modified Error Image									& \y & \n & \n & \y & \n & \n & \n \\
		Lim et al. 			& (2012) \cite{Lim2012}			& Combine spatial/temporal models						& \y & \n & \n & \y & \y & \y & \y \\
		Kim et al.			& (2013) \cite{Kim2013a}		& Optical flow clustering and Delaunay triangulation	& \y & \n & \n & \y & \n & \n & \n \\
		Yi et al. 			& (2013) \cite{Yi2013}			& Dual background model									& \y & \n & \y & \n & \n & \n & \n \\
		Lim and Han			& (2014) \cite{Lim2014}			& Superpixel segmentation								& \n & \y & \n & \y & \y & \y & \y \\
		Yun and Choi 		& (2015) \cite{Yun2015}			& Selectively update									& \y & \n & \y & \n & \y & \n & \n \\
		Chung et al. 		& (2016) \cite{Chung2016}		& Reduce background model errors		 				& \y & \n & \y & \n & \y & \y & \y \\
		Sun et al.			& (2016) \cite{Sun2016}			& Motion/appearance segmentations						& \y & \y & \n & \y & \n & \y & \n \\
		Makino et al.		& (2017) \cite{Makino2017}		& Score maps											& \y & \n & \y & \n & \n & \y & \n \\
		Wu et al.			& (2017) \cite{Wu2017}			& Coarse to fine strategy								& \y & \n & \n & \y & \n & \y & \n \\
		Yun et al.			& (2017) \cite{Yun2017}			& Slow moving objects									& \y & \n & \y & \n & \n & \n & \n \\
		\bottomrule
	\end{tabular}
	\caption{Split image in blocks methods summary. \textbf{RG}: Regular Grid, \textbf{S}: Superpixels, \textbf{MCI}: Motion Compensation on Image, \textbf{MCB}: Motion Compensation on Blocks, \textbf{BFM}: Background and Foreground Models, \textbf{MAM}: Motion and Appearance Models, \textbf{IM}: Iterative Method.}
\end{sidewaystable*}
%

\section{Datasets and evaluation metrics}
\label{sec:datasets_and_evaluation_metrics}
This section introduces the publicly available datasets and the quantitative evaluation metrics that be used on these datasets to measure the performance of a method and compare them.

%

\subsection{Existing datasets}
\begin{figure*}
	\begin{minipage}{\textwidth}
		\centering
		\includegraphics[width=0.2\textwidth]{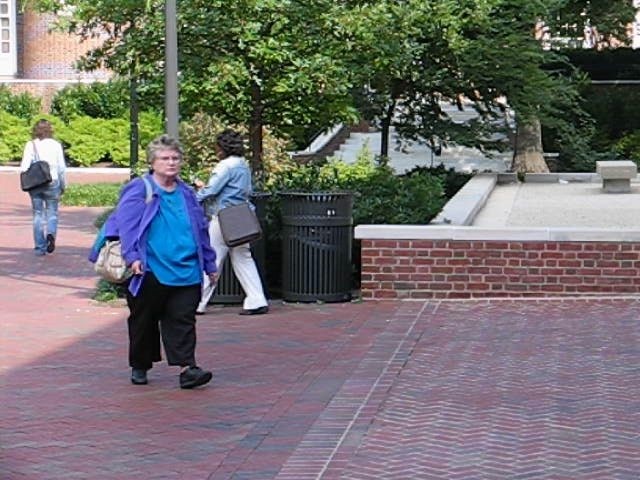}
		\includegraphics[width=0.2\textwidth]{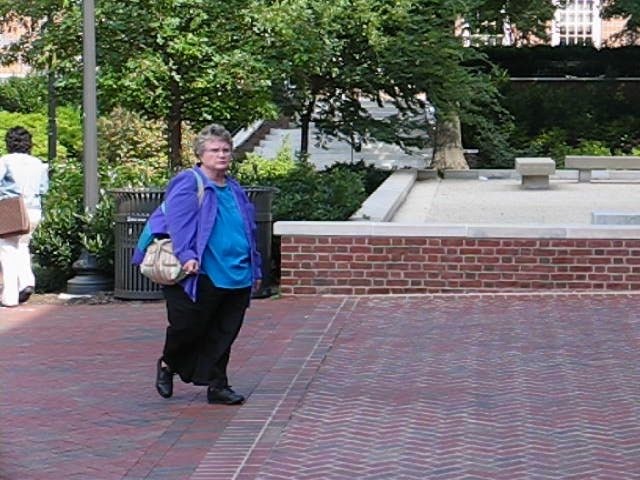}
		\includegraphics[width=0.2\textwidth]{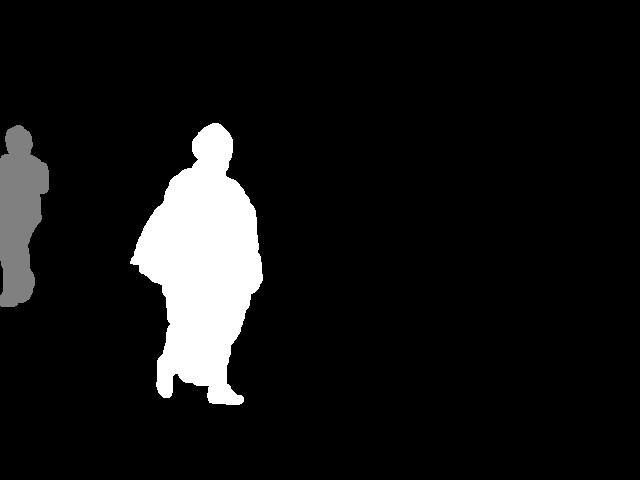}
		\subcaption{Hopkins 155+16 dataset, sequence people2}
	\end{minipage}
	\begin{minipage}{\textwidth}
		\centering
		\includegraphics[width=0.2\textwidth]{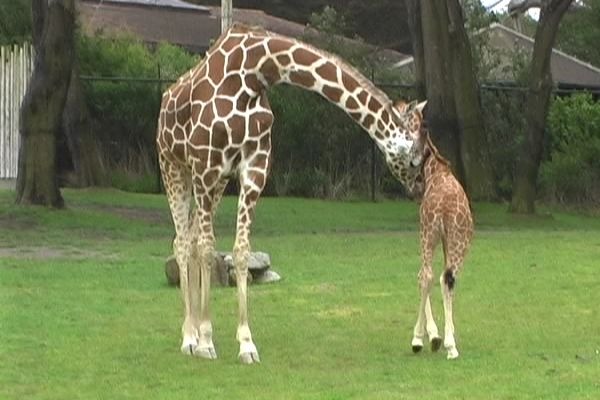}
		\includegraphics[width=0.2\textwidth]{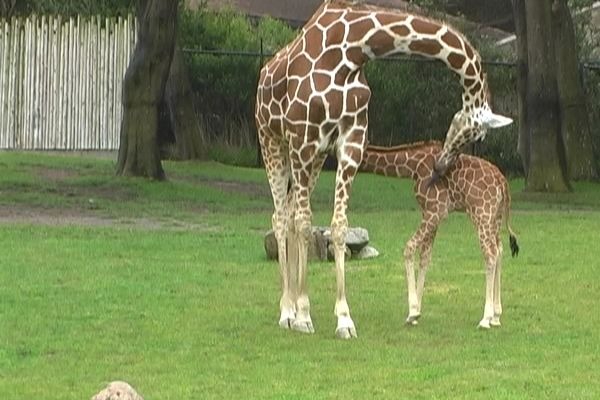}
		\includegraphics[width=0.2\textwidth]{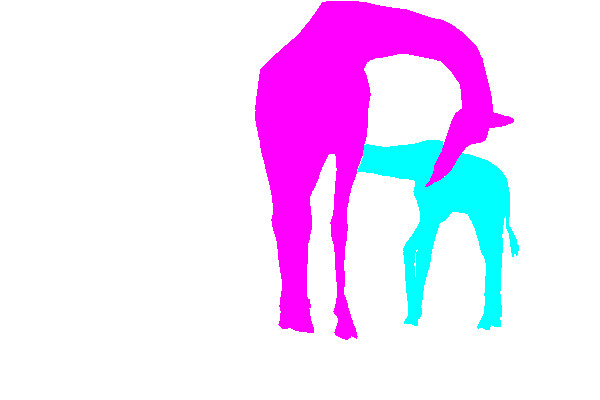}
		\subcaption{FBMS-59, sequence giraffes01}
	\end{minipage}
	\begin{minipage}{\textwidth}
		\centering
		\includegraphics[width=0.2\textwidth]{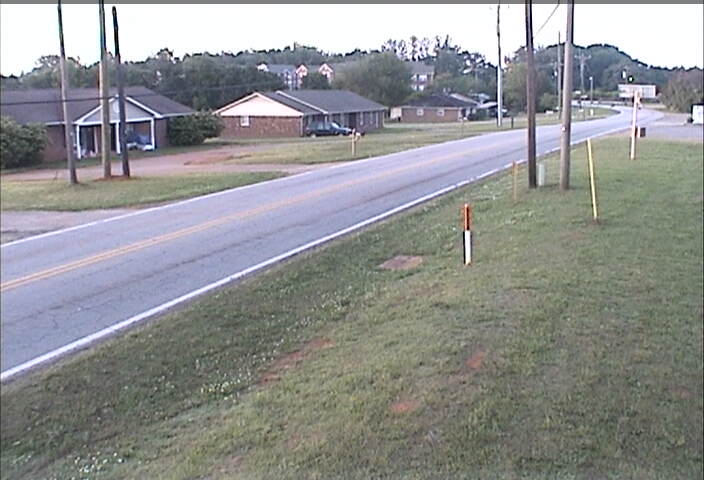}
		\includegraphics[width=0.2\textwidth]{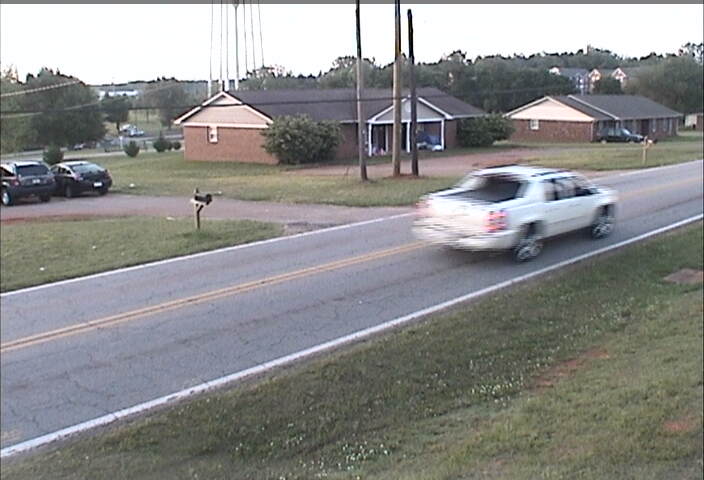}
		\includegraphics[width=0.2\textwidth]{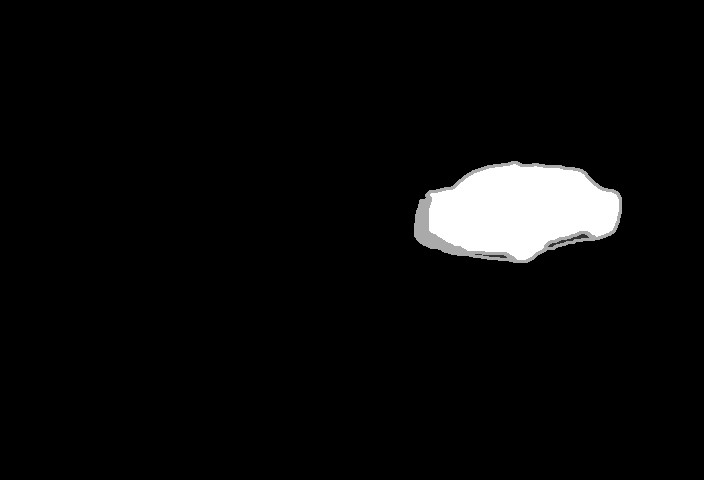}
		\subcaption{ChangeDetection.net, sequence continuousPan}
	\end{minipage}
	\begin{minipage}{\textwidth}
		\centering
		\includegraphics[width=0.2\textwidth]{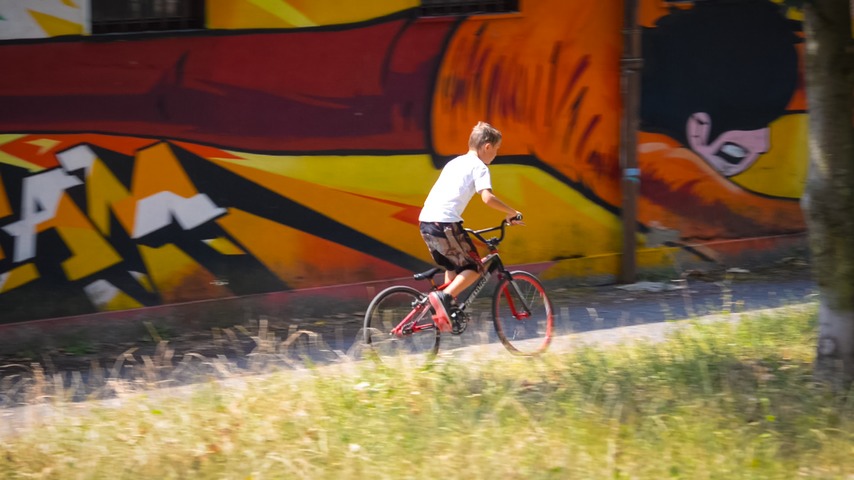}
		\includegraphics[width=0.2\textwidth]{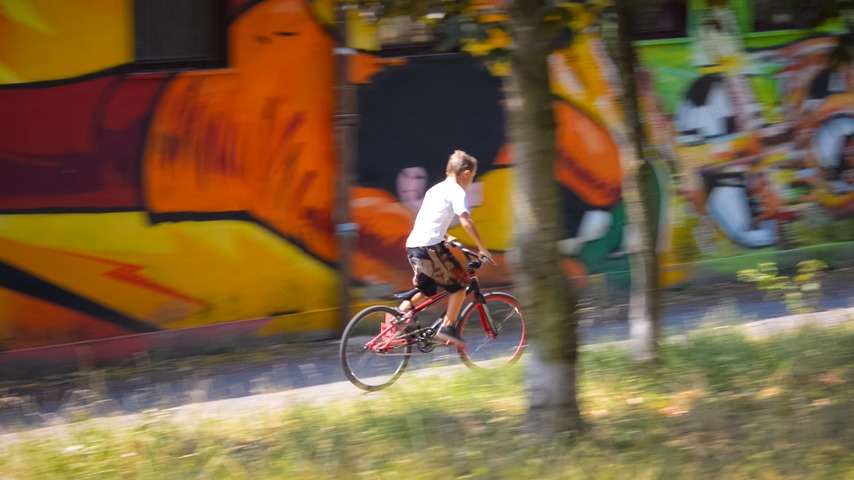}
		\includegraphics[width=0.2\textwidth]{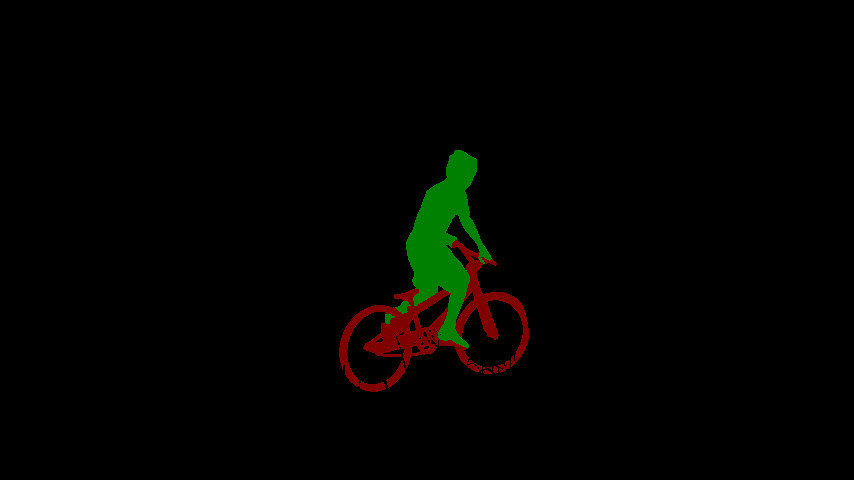}
		\subcaption{DAVIS, sequence bmx-trees}
	\end{minipage}
	\begin{minipage}{\textwidth}
		\centering
		\includegraphics[width=0.2\textwidth]{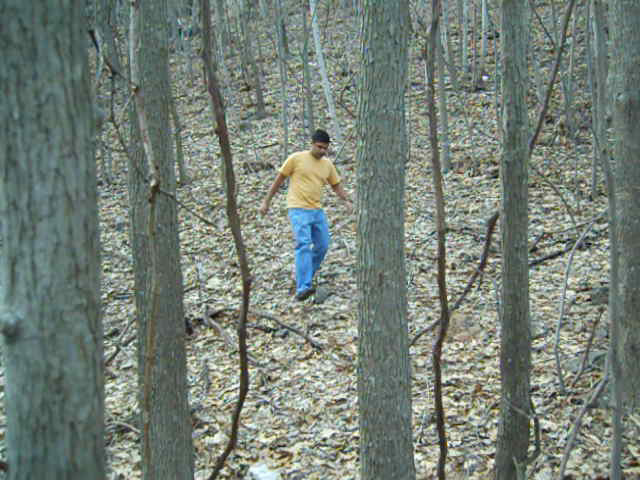}
		\includegraphics[width=0.2\textwidth]{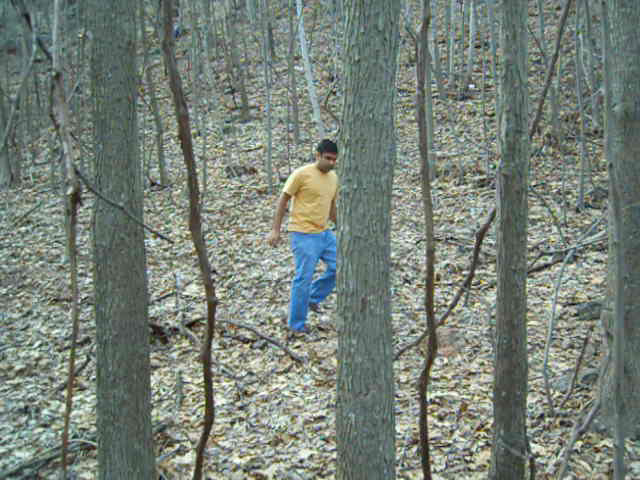}
		\includegraphics[width=0.2\textwidth]{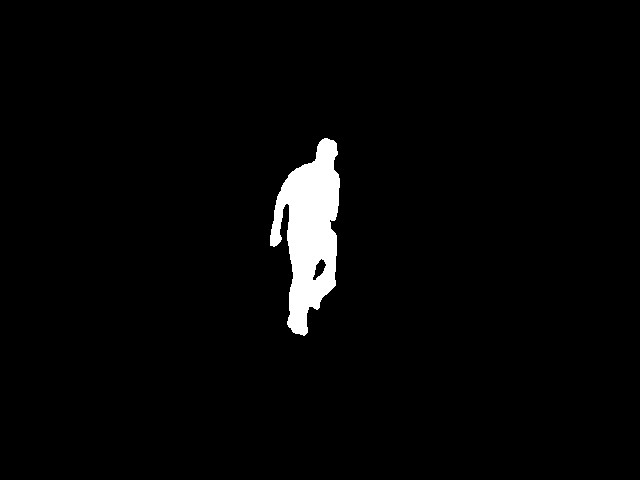}
		\subcaption{ComplexBackground, sequence forest}
	\end{minipage}
	\caption{Illustrations of datasets with input images and their ground truth. The two first columns are images taken by a moving camera and the third column is the ground truth of images from the second column.}
	\label{fig:illustration_datasets}
\end{figure*}

In order to test the performance of a moving object detection method with a moving camera, it is necessary to have video sequences whose each pixel of each frame are annotated. This section presents the datasets that can be used to evaluate and compare methods. Images and the ground truth taken from these datasets are presented in the figure~\ref{fig:illustration_datasets}. In the same way one writes this paper, only datasets that contain videos taken by a moving camera are presented.

\begin{itemize}
\item The \textbf{Hopkins 155+16} dataset was firstly introduced by Tron and Vidal \cite{Tron2007} and known as the Hopkins 155 dataset. This dataset was originally created to evaluate motion segmentation algorithms but the data can also be used for moving objects detection algorithms. There are 57 different videos, mostly taken by a moving camera and 114 sequences derived from these videos. The derived sequences differ from the original ones by their ground truth which represent a subset of motions in the video. For each sequence, complete trajectories of feature points and ground truth on the points are provided. For the 16 additional sequences, the trajectories contain missing data and outliers. Moving objects are chessboards in two-thirds of sequences and the last third contains cars and people.

\item \textbf{FBMS-59} dataset proposed by Ochs et al. \cite{Ochs2014} (Freiburg-Berkeley Motion Segmentation dataset) is an extension of the \textbf{BMS-26} dataset of Brox and Malik \cite{Brox2010a} (Berkeley Motion Segmentation dataset). The BMS-26 consists of 26 sequences, where 12 sequences come from the Hopkins 155 dataset, taken by a moving camera where most video sequences present high camera movements. Brox and Malik provided ground truth masks on some frames of the BMS-26 dataset, accumulating a total of 189 frames annotated. Annotations are masks where each moving object is pixel-accurate identified by a grayscale value. The FBMS-59 dataset extended the BMS-26 dataset with 33 additional video sequences with a total of 720 frames annotated. This dataset is decomposed into training and test sets. The masks provided can be easily used to evaluate moving objects detection algorithms. 

\item \textbf{ChangeDetection.net} called CDnet. There exist two versions of this dataset: CDnet 2012 \cite{Goyette2012} and CDnet 2014 \cite{Wang2014a}. Almost all sequences are taken by a static camera but in CDnet 2014, four sequences are taken by a PTZ camera. For each sequence, a ground truth mask is provided. The mask contains five labels: static, hard shadow, outside region of interest, unknown motion (usually around moving objects, due to semi-transparency and motion blur) and motion. Each label is associated to a gray color and a simple filter can be used on this mask to obtain a binary mask which can be used to evaluate a method.

\item The Densely Annotated VIdeo Segmentation \textbf{DAVIS} was proposed by Perazzi et al. \cite{Perazzi2016}. Three versions of the dataset were proposed: \cite{Perazzi2016}, \cite{Pont-Tuset2017}, \cite{Caelles2019}. The first version \cite{Perazzi2016} contains 50 different videos where only 5 videos were taking by a static or a shaking camera. For each video, a binary ground truth mask is given for each frame. In the two other versions of the dataset \cite{Pont-Tuset2017} and \cite{Caelles2019}, 40 videos were added. Among those 90 video sequences, only 10 were taking by a static or a shaking camera. In the same manner than for the first dataset version, for each frame of a video, a mask is given. The mask is not a binary mask but moving objects are classified into categories like human or bike according to colors. The background is still identified by the black color and it can be used to differentiate background from foreground.

\item \textbf{ComplexBackground} is a dataset proposed by Narayana et al. \cite{Narayana2013} and contains five video sequences taken by a hand-held camera. Each video contains 30 frames and 7 frames are used for the ground truth as a binary mask. These videos contain one or several moving objects and the static scene presents significant depth variations.
\end{itemize}

%
\subsection{Evaluation metrics}
Thanks to the publicly available datasets and their associated ground truth, quantitative metrics are used to evaluate the performance of background/foreground segmentation approaches and compare them together.

\bigskip

According to the ground truth, the pixel are categorized into one of these four categories:
\begin{itemize}
\item True Positive (TP): the number of pixel correctly labeled as foreground. Also known as \emph{hit}.
\item True Negative (TN): the number of pixel correctly labeled as background. Also known as \emph{correct rejection}.
\item False Positive (FP): the number of pixel incorrectly labeled as foreground. Also known as \emph{false alarm} or \emph{Type I error}.
\item False Negative (FN): the number of pixel incorrectly labeled as background. Also known as \emph{miss} or \emph{Type II error}.
\end{itemize}

\begin{figure*}[htbp]
	\begin{minipage}{\textwidth}
		\centering
		\includegraphics[width=0.3\textwidth]{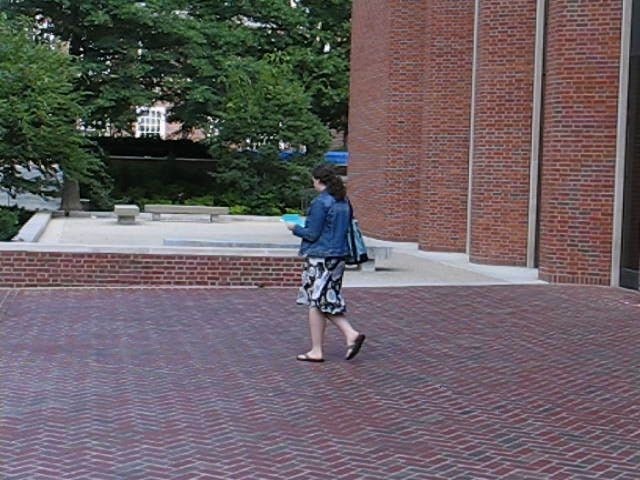}
		\includegraphics[width=0.3\textwidth]{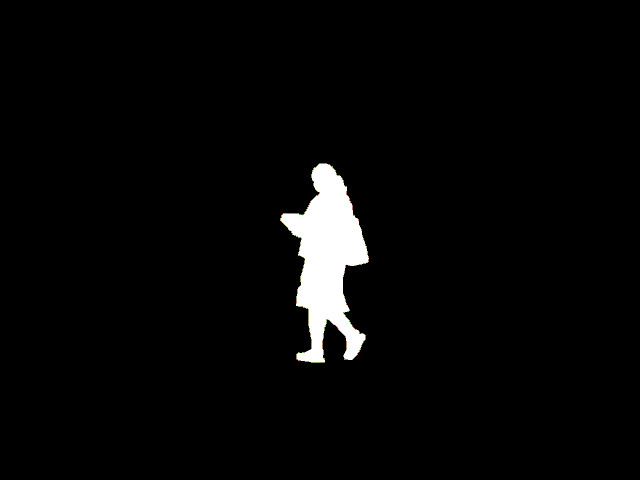}
		\includegraphics[width=0.3\textwidth]{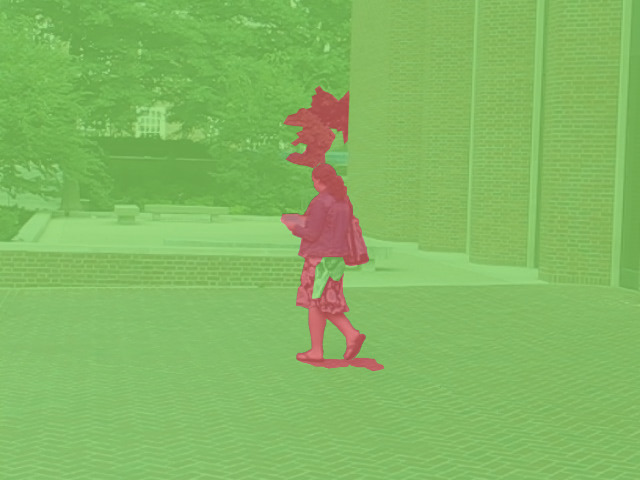}
	\end{minipage}
	\caption{An example of background/foreground segmentation on the people01 sequence the from the Hopkins dataset. Left: the original image. Center: the ground truth. Right: an example of background/foreground segmentation where green pixels are labeled as background and red pixels are labeled as foreground.}
	\label{fig:evaluation_bg_segmentation_example}
\end{figure*}

Three measures are commonly used to evaluate background subtraction algorithms: the precision, the recall and the F-score.

\begin{itemize}
\item The \emph{precision} (also known as \emph{positive predictive value}) is the proportion of pixels that are correctly detected as moving among all pixels detected as moving by the algorithm.
\begin{equation}
Precision = \frac{TP}{TP+FP}
\end{equation}

\item The \emph{recall} (also known as \emph{sensitivity}, \emph{hit rate} or \emph{true positive rate}) is the proportion of pixels that are correctly detected as moving among all pixels that belong to moving objects in the ground truth.
\begin{equation}
Recall = \frac{TP}{TP+FN}
\end{equation}
\item The \emph{F-score} (also known as \emph{F1 score} or \emph{F-measure}) is the combination of precision and recall. It is the harmonic mean of precision and recall measures:
\begin{equation}
F-score = \frac{2 \times Recall \times Precision}{Recall+Precision}
\end{equation}
\end{itemize}

Several other measure metrics are also used:

\begin{itemize}
\item The \emph{Accuracy} is the proportion of pixels detected as moving among all the labeled pixels.
\begin{equation}
Accuracy = \frac{TP + TN}{TP + TN + FP + FN}
\end{equation}

\item The \emph{Specificity} (also known as \emph{selectively} or \emph{true negative rate}) is the proportion of pixels that are correctly detected as static among all pixels that belong to static objects in the ground truth.
\begin{equation}
Specificity = TN / (TN + FP)
\end{equation}

\item The \emph{false positive rate} (also known as \emph{fall-out}) is the proportion of pixels that are incorrectly detected as moving among all pixels that belong to static objects in the ground truth.
\begin{equation}
False Positive Rate : FP / (FP + TN)
\end{equation}

\item The \emph{false negative rate} (also known as \emph{miss rate}) is the proportion of pixels that are incorrectly static as moving among all pixels that belong to moving objects in the ground truth.
\begin{equation}
False Negative Rate : FN / (TP + FN)
\end{equation}
\end{itemize}
%
\section{Conclusion}
We have proposed in this paper a review of methods for moving objects detection with a moving camera categorized into eight different approach groups divided into two big categories. We have chosen to separate the methods into these two categories, one plane and several planes, since the approach to use depends on the scene configuration. For each group, the following conclusions can be made:
\begin{itemize}
\item For the approaches based \textbf{panoramic background subtraction}, a panorama of the observed scene is first constructed. Then, the current image is registered to the background model in order to do the subtraction and obtain the moving objects. These approaches are often used in the context of video surveillance with a PTZ camera. The \emph{panoramic background subtraction} approach is well suited for this kind of camera since the part of the scene that the camera can observed is limited because it cannot perform a translation. A special attention must be paid on the construction of the panorama because errors can be accumulated and caused errors in the background subtraction step.
\item When several cameras are used, static and moving, it could be interesting to couple the information to detect moving objects. In the \textbf{dual cameras} approaches, when a moving object is detected in the static camera, generally with a large-view, the moving camera, generally a PTZ camera, will move to detect the moving object. The advantage of using the large-view image, compared to the panoramic background subtraction, is that the whole background model is updated with the new frames.
\item The background subtraction with a \textbf{motion compensation} approach is the most popular in the literature, as shown by the table~\ref{tab:motion_compensation_methods_summary}. The two advantages of this method are the ease of implementation and its low time computation. The compensation is a 2D transformation which approximates the scene by a plane. When the parallax is small, it can be handle after the compensation but when the parallax is too large, this approach cannot be used.
\item Contrary to the three previous approaches, the motion of the camera is not compensated to compute the background subtraction. The \textbf{subspace segmentation} approach is based on the apparent motion, computed by optical flow algorithms on feature points or on the entire image. These trajectories are then clustered or segmented into a subspace representation. The clusters segmented as background generally reflect a plane in the scene.
\item The \emph{motion segmentation} approach is also based on trajectories. The motions are analyzed and segmented according to their similarities. The methods presented in this survey go further than just segmented the motions, a background or foreground label is associated to the sets of motions. In the same manner as the subspace representation approach, the background motions generally reflect a plane in the scene.
\item  The \textbf{Plane+Parallax} approach was not much studied in the context of detecting moving objects. To the best of our knowledge, only three different methods relate about the \emph{Plane+Parallax} decomposition. This scene representation performs well when the scene contains few parallax and difficulties arise when the scene is composed of several planes.
\item  In the \textbf{multi planes} approaches, the scene is approximated by several planes, reals or not. With such a scene representation, most of the parallax effect is directly handled. Nevertheless, if a moving object is big enough in the images, it can be approximated by a plane and considered as a part of the background.
\item  Rather than representing the scene by several planes, the \textbf{split image in blocks} approaches divided the image into several blocks. Each block is processed individually in order to find the foreground objects. As in the multi planes approaches, a moving object as to be small in a block in order to approximate the block as a plane.
\end{itemize}
Among all the challenges presented in the section~\ref{sec:Challenges}, the \emph{Moving Camera} and the \emph{Motion Parallax} are usually the main contributions of the papers that use a moving camera. The other challenges are generally overcome by using or adapting solutions which come from methods with a static camera. Approaches designed for one plane are well suited for the scenes that can be approximated by one plane with few parallax whereas the approaches in the several parts categories can handle more parallax. In both cases, the methods make the assumption that the apparent motion of the scene is pretty the same while in some configuration scene and camera motion, the scene can appears in the images with different motions. It could be interesting to investigate and propose methods that address this case. Moreover, the number of datasets which contain this kind of video is quite small. Narayana et al. \cite{Narayana2013} propose videos with complex background in their \textbf{ComplexBackground} dataset, but only five videos are provided. In the same manner, some challenges datasets are missing as underwater videos taken by a moving camera. With the recent advances in Deep Learning, it could be interesting to test different architectures on the problem of moving objects detection with a moving camera, as the combination of an appearance network and a motion network \cite{Jain2017, Tokmakov2017} or a network which reconstructs the background from an image \cite{Minematsu2018}.

	\section*{Acknowledgments}
	This research did not receive any specific grant from funding agencies in the public, commercial, or not-for-profit sectors.
	
	\bibliography{Survey,StaticCamera,MovingCamera}

\end{document}